\documentclass[11pt]{article}

\usepackage[preprint]{acl}

\usepackage{times}
\usepackage{latexsym}
\usepackage[T1]{fontenc}
\usepackage[utf8]{inputenc}
\usepackage{microtype}
\usepackage{inconsolata}
\usepackage{graphicx}
\usepackage{amsmath}
\usepackage{booktabs}
\usepackage{array}
\usepackage{tabularx}
\usepackage[table]{xcolor}
\usepackage{placeins}
\usepackage{float}
\usepackage{tikz}
\usetikzlibrary{arrows.meta,positioning}

\newcommand{\method}{SCIT}
\newcommand{\twostep}[2]{#1/#2}

\newcommand{\appendixtablefont}{\scriptsize}
\newcolumntype{Y}{>{\raggedright\arraybackslash}X}
\newcolumntype{L}[1]{>{\raggedright\arraybackslash}p{#1}}
\definecolor{latentgreen}{HTML}{14824F}
\definecolor{prefixblue}{HTML}{2563EB}
\definecolor{mixedviolet}{HTML}{6D4AFF}
\definecolor{weakamber}{HTML}{B7791F}
\definecolor{pathviolet}{HTML}{7C3AED}
\definecolor{gatebg}{HTML}{EEF2FF}
\definecolor{suffbg}{HTML}{ECFDF5}
\definecolor{necbg}{HTML}{FFF7ED}
\definecolor{decodebg}{HTML}{EFF6FF}
\definecolor{riskbg}{HTML}{F5F3FF}
\newcommand{\latentverdict}{\textcolor{latentgreen}{\textbf{LATENT-TAIL}}}
\newcommand{\prefixverdict}{\textcolor{prefixblue}{\textbf{PREFIX SHIFT}}}
\newcommand{\boundaryverdict}{\textcolor{weakamber}{\textbf{NO CALL}}}
\newcommand{\pathverdict}{\textcolor{pathviolet}{\textbf{PATH-SENSITIVE}}}
\newcommand{\auditverdict}{\textbf{AUDITED}}

\title{\method{}: Testing Causal Cache Carriers in Latent Chain-of-Thought Models}

\author{
  \href{https://openreview.net/profile?id=~YI_DING21}{Yi Ding}
  \quad
  \href{https://openreview.net/profile?id=~Lijun_Huang1}{Lijun Huang}
  \quad
  \href{https://openreview.net/profile?id=~Menglin_Yang3}{Menglin Yang}
  \\
  The Hong Kong University of Science and Technology (Guangzhou) \\
  Guangzhou, China
}

\begin{document}
\maketitle

\begin{abstract}
Latent chain-of-thought models move intermediate reasoning from emitted text into continuous states, improving compactness but hiding the causal object. We introduce SCIT, the Suffix Cache Interchange Test, a causal protocol that constructs exact source-recipient counterfactuals, patches declared cache segments, and identifies which transformer object carries the counterfactual computation. SCIT combines sufficiency tests with K/V component splits, hidden-state controls, semantic source controls, decoded validation, and matched corruption. On CODI-GPT2 and a Sim-CoT-style GPT-2 reproduction, counterfactual arithmetic transfers primarily through value-cache suffix trajectories rather than hidden states, keys, reusable answer slots, or single-token triggers. Complete sufficiency-and-necessity evidence for the late-value-suffix mechanism holds for the main CODI-GPT2 checkpoint; the Sim-CoT-style checkpoint shows the same sufficiency and decoded-control pattern but insufficient matched-corruption evidence for a necessity call. Beyond these local arithmetic cells, SCIT reveals carrier-regime shifts: arithmetic-like GPT-2/1B cells preserve latent-tail value/KV transfer, whereas competent 8B and repaired non-arithmetic cells route through prompt-prefix or full-cache K/V; boundary cells receive no mechanism call. SCIT therefore contributes a cache-level diagnostic, a checkpoint-specific GPT-2 arithmetic mechanism, and a competence-gated carrier map rather than a universal latent-tail claim.
\end{abstract}

\section{Introduction}

Chain-of-thought prompting made intermediate reasoning visible as text, but latent and implicit chain-of-thought methods move part of that work into continuous internal states \citep{wei2022chainofthought,hao2024training,shen2025codi,wei2025simcot}. This shift is attractive because it can reduce long scratchpads and expose a compact reasoning interface. It also changes the interpretability target: if the model no longer writes a rationale, faithfulness cannot be checked by reading one. We need to ask where the hidden computation is implemented inside the transformer.

Existing latent-CoT analyses often treat the latent step itself as the causal variable, asking whether perturbing or replacing that step changes the answer \citep{li2026dynamics}. Such analyses identify when latent computation matters, but not which transformer object carries the effect. A positive latent-step intervention could be explained by the current hidden vector, key-side routing, value-side content, a broad cache history, or a source that happens to share the target answer. These hypotheses are not logically exclusive: a value trajectory can encode a reusable feature. The empirical question is which object is causally sufficient and which simpler source or component controls can match the oracle effect.

We introduce \method{}, the Suffix Cache Interchange Test, to turn this ambiguity into a carrier test. \method{} constructs exact source--recipient counterfactuals, patches declared cache segments during the latent rollout, and scores whether the recipient moves toward the source-defined counterfactual answer. The test is deliberately grid-like: it varies cache segment, K/V component, hidden-state coupling, source semantics, decoded behavior, and matched corruption. The output is therefore a carrier map rather than a single positive patch score.

We instantiate \method{} with controlled arithmetic counterfactuals whose recipient, counterfactual, semantic-control, and random answers are known exactly. This synthetic setting is a causal-identification device, not a benchmark-breadth claim: open-ended datasets rarely provide exact source--recipient counterfactuals, partial-variable controls, and known negative labels without extra assumptions.

Our claims are deliberately tiered. The \emph{method} claim is that \method{} converts latent-step intervention success into a structured cache-carrier diagnosis. Complete sufficiency-and-necessity evidence for the \emph{late-value-suffix mechanism} is established in a single checkpoint---the main CODI-GPT2 arithmetic checkpoint evaluated here: counterfactual answers transfer through value-cache suffixes, with the strongest restricted effect in middle-to-late values, and the same region passes matched-corruption necessity; hidden-only, key-only, one-token, same-answer, and partial-variable controls are much weaker. The Sim-CoT-style checkpoint shows the same sufficiency and decoded-transfer pattern, but its matched-corruption effect is insufficient for a full necessity call. The \emph{generalization} claim is diagnostic rather than universal: arithmetic-like GPT-2/1B cells preserve latent-tail value/KV transfer, while several competent 8B or repaired non-arithmetic cells shift to prompt-prefix or full-cache K/V. Different-segment sufficiency--corruption agreement is reported as a carrier shift, not as replication of the CODI-GPT2 mechanism, and no-call boundary cells receive no mechanism call.

We use three status terms throughout. An \emph{active carrier} is the smallest gated cache segment that is sufficient under oracle interchange and not matched by the main disjoint competitor. A \emph{carrier shift} is a gated cell where transfer exists but the active segment differs from the GPT-2 arithmetic latent-tail carrier. A \emph{no-call} row is a cell where competence, decoded behavior, support, or source-control checks prevent a mechanism interpretation.

The paper therefore contributes three linked but separable results. First, it introduces \method{} as a cache-level causal contract for latent-CoT rollouts. Second, it identifies a checkpoint-specific CODI-GPT2 arithmetic carrier: a context-bound value-cache suffix trajectory, with sufficiency-only corroboration from the Sim-CoT-style checkpoint. Third, it reports a competence-gated carrier-regime map that separates local mechanism evidence from scale, task, seed, and support boundaries. Appendix Tables~\ref{tab:intervention-defs}--\ref{tab:experiment-battery} give full intervention, prompt, and battery details.

The rest of the paper is organized as follows: Section~2 situates SCIT relative to latent-CoT and mechanistic-interpretability work; Section~3 defines the protocol; Section~4 describes the checkpoints, templates, and evaluation battery; Section~5 presents the local mechanism and carrier-regime results; Section~6 discusses scope and future directions; Section~7 concludes; and the appendix gives the full evidence ledger.

\section{Related Work}

\paragraph{Explicit reasoning traces and faithfulness.}
Scratchpads and chain-of-thought prompting made intermediate reasoning a core interface for multi-step computation \citep{nye2021scratchpads,wei2022chainofthought}. Follow-up prompting methods showed that explicit traces can be elicited, sampled, or decomposed in several ways, including zero-shot CoT, self-consistency, and least-to-most prompting \citep{kojima2022zeroshot,wang2022selfconsistency,zhou2022leasttomost}. But a readable rationale is not automatically a faithful explanation: NLP faithfulness work separates plausible explanations from causal ones, attention weights need not explain decisions, and CoT text can rationalize biased outputs or omit cues that the model actually used \citep{jacovi2020faithfully,jain2019attention,turpin2023language,lanham2023measuring,chen2025reasoning}. \method{} follows this causal view of faithfulness: the object of study is not whether a trace is readable, but whether an internal object is sufficient and selectively required for a counterfactual computation.

\paragraph{Latent and implicit chain-of-thought.}
Latent or implicit CoT methods move part of the reasoning process into continuous internal states \citep{hao2024training,shen2025codi,wei2025simcot}. Coconut feeds hidden reasoning states back as continuous inputs, while CODI and Sim-CoT compress, distill, or supervise latent steps to improve efficiency, stability, or interpretability. Recent causal analyses of latent-CoT dynamics treat latent steps as manipulable variables and study step-wise necessity, influence propagation, and representational commitment \citep{li2026dynamics}. Our work is complementary: rather than treating the latent step as the causal variable, we ask which specific transformer object (e.g., key cache, value cache, hidden state, or cache segment) implements the causal effect.

\paragraph{Reading hidden states versus causal localization.}
Probing and decoding methods such as the Tuned Lens and Patchscopes make hidden representations more inspectable by translating internal states into token distributions or natural-language explanations \citep{belrose2023tunedlens,ghandeharioun2024patchscopes}. These methods are complementary to \method{}: they can show that information is recoverable from a state, but recoverability does not by itself show that the model used that state in the computation. \method{} therefore treats decoding as a supporting check and relies on interchange, component controls, semantic source controls, and matched corruption for the main causal claim.

\begin{figure*}[!t]
\centering
\includegraphics[width=0.98\textwidth]{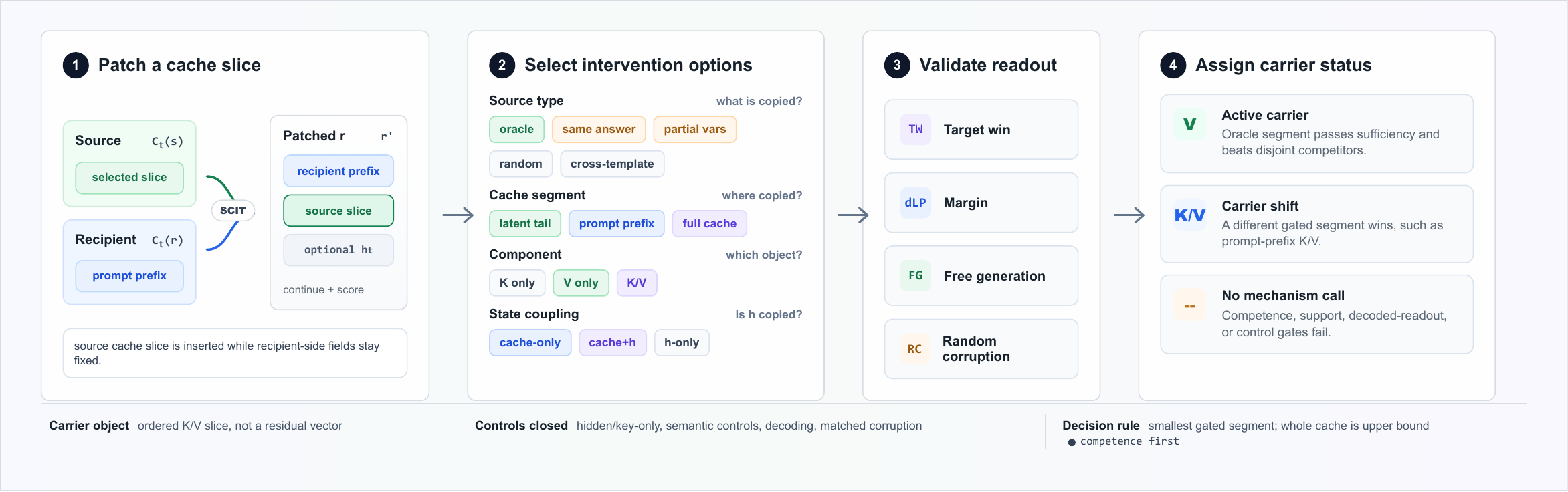}
\caption{\method{} carrier-test protocol. A source cache slice and recipient-side fields are combined into a patched recipient rollout, then evaluated under a fixed intervention grid. The patched rollout is validated with target-win, margin, free-generation, and matched-corruption readouts before assigning an active-carrier, carrier-shift, or no-call status. Segment calls use the smallest gated disjoint segment; whole-cache replacement is an upper-bound control. TW: target win; dLP: log-probability drop.}
\label{fig:scit}
\end{figure*}

\paragraph{Causal interventions and circuit discovery.}
\method{} sits in the mechanistic-interpretability lineage of causal abstraction, interchange intervention, causal tracing, activation patching, path/circuit analysis, and automated circuit discovery \citep{geiger2021causal,geiger2022inducing,meng2022locating,elhage2021mathematical,olsson2022induction,wang2023interpretability,conmy2023acdc,heimersheim2024activation,zhang2024best}. This line of work connects localized interventions to attention heads, circuits, factual associations, and circuit-discovery pipelines. The novelty in \method{} is not counterfactual interchange alone; interchange intervention accuracy (IIA) already provides that logic when the causal variable is pre-specified. SCIT instead uses the intervention grid to discover which transformer object should count as the carrier in the first place: cache segment, K/V component, hidden-state coupling, source semantics, decoded behavior, and matched corruption are all varied together to separate a carrier map from a single circuit or a benchmark score.

\section{Method}
\label{sec:method}

Figure~\ref{fig:scit} summarizes \method{} as a four-step carrier-test contract. The first step defines the patch object, the second fixes the intervention grid, the third validates the readout, and the fourth applies the carrier decision rule. A mechanism call is made only after the carrier object is specified, simpler control explanations are closed, and competence gating is applied.

\subsection{SCIT protocol}

For intuition, consider a recipient problem in which Alice has 7 red and 8 blue balls and gives away 2 red balls, yielding 13. A valid source problem supplies a counterfactual post-transfer red count of 11. Combining that source quantity with the recipient's 8 blue balls defines the verified counterfactual answer 19. \method{} asks which internal memory region must be transplanted for the recipient model to change from 13 to 19.

\paragraph{Step 1: patch a cache slice.}
Consider a latent-CoT model that performs $T$ latent reasoning steps before scoring an answer. At latent step $t$, a rollout has a hidden state $h_t$ and a transformer cache $C_t=\{(K_t^\ell,V_t^\ell)\}_{\ell=1}^L$. Here $C_t$ denotes the K/V entries available at the patch point, including entries written by latent positions through step $t$, while $h_t$ is the current residual state used to continue the rollout or read out the answer. Thus $h_t$ is not part of the cache object unless an intervention explicitly copies it.

For a recipient problem $r$ and a source problem $s$, \method{} runs both rollouts to step $t$, inserts a selected slice from $C_t(s)$ into the corresponding positions of $C_t(r)$, optionally copies $h_t(s)$, and then continues the recipient rollout as the patched state $r'$. The primary arithmetic hypothesis is that the latent-tail suffix written during implicit reasoning carries the computation; in scope-calibration cells, the same interchange contract also tests prompt-prefix and whole-cache replacement to ask whether the carrier has moved.

Operationally, a \emph{trajectory} is the ordered collection of K/V entries written by the rollout over a suffix of cache positions, restricted to the tested layer/head region when a component scan is used. A value-cache trajectory is therefore not a single residual-stream direction, attention pattern, or decoded symbolic variable; it is a causal intervention object whose sufficiency is tested by replacing that ordered value-cache slice.

\paragraph{Variable-length rollouts.}
When source and recipient rollouts stop at different realized steps $(T_r,T_s)$, \method{} aligns latent-cache positions by reverse distance from their respective stops rather than by absolute position. Let $c_x=T_x-1$. For $k\leq\min(c_r,c_s)$, we patch $\mathrm{KV}^{\mathrm{lat}}_r[c_r-k:c_r]$ with $\mathrm{KV}^{\mathrm{lat}}_s[c_s-k:c_s]$, retain unmatched recipient positions, and treat the current hidden state $h$ as a separate intervention object. Carrier calls under variable stopping are therefore conditioned on the source and recipient stopping policy.

\paragraph{Step 2: select intervention options.}
Each \method{} run fixes four choices shown in Figure~\ref{fig:scit}: source type, cache segment, K/V component, and hidden-state coupling. Cache segments include latent tail, prompt prefix, and full cache; full-cache replacement is retained as an upper-bound control rather than a minimal-carrier verdict. Component choices include K-only, V-only, and natural K/V patching; hidden-state coupling compares cache-only, cache+h, and h-only variants. For GPT-2 layer localization, we use all layers, a predeclared final-third group (layers 8--11), its complement (0--7), and two diagnostic sub-blocks (8--9 and 10--11); Appendix Table~\ref{tab:gpt2-block-scan} scans contiguous layer blocks behind the middle-to-late localization claim. For trajectory checks, we also vary suffix length to test whether the effect is carried by the final cache token alone. Appendix Table~\ref{tab:intervention-defs} defines the full grid.

Key-only and value-only patches are intentionally surgical interventions, not states that a clean transformer forward pass would normally produce. They splice source keys with recipient values, or source values with recipient keys, to test component sufficiency under a fixed downstream decoder. We therefore interpret them as causal diagnostics rather than as samples from the model's natural activation distribution: a successful value-only patch shows that source value content is sufficient under recipient routing, while a failed key-only patch rules out key-only sufficiency but does not claim that keys are irrelevant in every natural computation. We pair these artificial splits with natural K/V patches, hidden-state controls, decoded behavior, and matched corruption before making a mechanism call.

The arithmetic sources are synthetic by design: \method{} needs exact source--recipient counterfactual answers, known intermediate variables, and negative semantic controls. Balls, books, and coins are lexical variants of one two-hop arithmetic program, so they test surface-wording robustness rather than task diversity; Appendix Table~\ref{tab:prompt-templates} gives the full prompts. Each example supplies a recipient answer $a_r$, an oracle counterfactual answer $a_{cf}$, random controls, and semantic controls that preserve only selected information. Same-answer, same-intermediate, partial-variable, and cross-template partial sources preserve one candidate feature while breaking the full counterfactual computation; exact cross-template sources preserve the full arithmetic state in another lexical family. We call the arithmetic trajectory \emph{context-bound} only if exact counterfactual sources transfer while same-answer, partial-variable, one-token, hidden-only, and key-only controls do not match the oracle effect.

\paragraph{Step 3: validate the readout.}
The patched recipient rollout is continued and scored against the same candidate-answer set as the clean rollout. The primary metric is \emph{target win}: whether the expected answer has the highest summed teacher-forced log-probability among the fixed candidate answers. Because target win is a closed-set argmax, we also track a patched-minus-clean counterfactual-vs-recipient log-probability margin,
\[
\begin{aligned}
\Delta_{\mathrm{cf-rec}} =
&[\log P(a_{cf})-\log P(a_r)]_{\mathrm{patched}} \\
&-
[\log P(a_{cf})-\log P(a_r)]_{\mathrm{clean}} ,
\end{aligned}
\]
with both probabilities evaluated under the same teacher-forced answer string. Positive $\Delta_{\mathrm{cf-rec}}$ means the patch increases relative evidence for the counterfactual answer beyond the clean recipient baseline. Four-answer cells have nominal chance near $1/4$, but semantic-control interpretation relies on oracle-control margins, decoded behavior, and tie audits rather than chance alone. For headline semantic controls, we also continue the patched state with greedy or sampled decoding and parse whether the free-generation answer matches the counterfactual. Appendix~\ref{app:method-setup} gives the candidate sets and row-specific margin use, and Appendix Table~\ref{tab:metric-audit} reports exact and near ties.

Sufficiency alone does not show that a region is required, so the validation step also includes matched random corruption. We replace the recipient's selected cache entries with entries from a same-template, same-range random source and ask whether the recipient answer $a_r$ remains the winner. This is less out-of-distribution than zero ablation but conservative, because another arithmetic rollout can share generic task structure. Unless a table states otherwise, each recipient receives one matched random source; Appendix~\ref{app:method-setup} gives the matching rule, and we report both target win and recipient-answer log-probability drop.

\begin{table*}[t]
\centering
\begingroup
\scriptsize
\renewcommand{\arraystretch}{1.0}
\setlength{\tabcolsep}{3pt}
\begin{tabularx}{\textwidth}{@{}L{0.16\textwidth}L{0.12\textwidth}Y L{0.16\textwidth}@{}}
\toprule
\textbf{Claim} & \textbf{Verdict} & \textbf{Key evidence and risk closed} & \textbf{Where} \\
\midrule
Local arithmetic carrier & \latentverdict{} & GPT-2 arithmetic passes competence; value suffixes transfer (all/late V $0.996$--$1.000$, layer 8--9 V $0.875$--$0.992$), while key/$h$ and semantic controls are weak. CODI corruption closes necessity (TW $0.211$); Sim-CoT is sufficiency-only/directional. & Tables~\ref{tab:localization}, \ref{tab:semantic}, \ref{tab:necessity}, \ref{tab:freegen} \\
Arithmetic scale check & \latentverdict{} & 1B arithmetic-like cells preserve the active latent carrier (active $1.00$ vs inactive $0.00$; corruption active TW $0.00$ vs inactive $1.00$). The same arithmetic-style diagnostic shifts at 8B. & Fig.~\ref{fig:carrier-regime-main}; App. Tables~\ref{tab:carrier-evidence-ledger}, \ref{tab:carrier-split} \\
8B field-routed cells & \prefixverdict{} & Competent 8B arithmetic/entity/relation/expression cells route through prompt-prefix/full K/V: direct-pre sufficiency is already $.969$--$1.000$, latent $0.00$, active-corruption dLP $13.64$--$21.07$. In arithmetic/expression/list/relation follow-ups, post-prefix zero and field ablations collapse readout from visible fields while latent-suffix ablation does not. & Fig.~\ref{fig:carrier-regime-main}; Table~\ref{tab:field-routing-main}; App. Tables~\ref{tab:main-8b-time-prefix}, \ref{tab:8b-no-latent}, \ref{tab:8b-prefix-readout-ablation}, \ref{tab:field-routing-hardening}, \ref{tab:repair-boundary} \\
Boundary discipline & \boundaryverdict{} & Qwen3-4B stays near chance ($0.25$--$0.33$); graph/finite-state/OOD/support-limited cells fail competence, support, or decoded-readout gates. No carrier is interpreted below gate. & Tables~\ref{tab:carrier-evidence-ledger}, \ref{tab:repair-boundary}, \ref{tab:fixedrule-support-boundary} \\
Training-path sensitivity & \pathverdict{} & T16 and new CODI seed audits pass clean gates but enter different basins; new CODI seeds show only $0.049$--$0.138$ late-8--9 value sufficiency and $0.026$--$0.089$ corruption drops. & Tables~\ref{tab:training-horizon-boundary}, \ref{tab:codi-multiseed-bootstrap} \\
Metric robustness & \auditverdict{} & Competence sweep $.70/.80/.90$ is stable; 16 carrier-threshold settings and 12/12 large-$N$ calls are unchanged; exact/near ties are rare ($0.063\%/0.072\%$). & Table~\ref{tab:metric-audit} \\
\bottomrule
\end{tabularx}
\caption{Compressed claim dashboard. Each row gives the mechanism verdict, the shortest evidence bundle needed to interpret it, and the numerical source; Appendix Table~\ref{tab:carrier-evidence-ledger} gives the full gate/sufficiency/corruption/decoded ledger.}
\label{tab:headline-results}
\endgroup
\end{table*}

\paragraph{Step 4: assign carrier status.}
Putting these pieces together, we interpret each cell with a fixed decision rule. Mechanistic interpretation first requires clean task competence: the unpatched recipient and the corresponding clean source/oracle prompts must reach at least $0.80$ parsed free-generation accuracy, or at least $0.80$ closed-set accuracy in cells explicitly marked as teacher-forced competence checks. Among gated cells, we call a segment an active sufficiency carrier only when oracle interchange on that segment reaches at least $0.80$ target win, the main competing disjoint segment is at most $0.25$, and decoded behavior agrees for headline rows. Whole-cache K/V is treated as an upper-bound control rather than a competing minimal segment: when both whole-cache and a disjoint segment pass, the carrier verdict is assigned to the smallest passing segment. These are reporting thresholds, not fitted estimators; Appendix Table~\ref{tab:metric-audit} sweeps competence and carrier thresholds and shows that the large-$N$ 1B/8B carrier calls are unchanged.

We reserve \emph{full necessity} for rows where the same segment that is sufficient under oracle interchange is also selectively required under matched corruption: corrupting that active segment must reduce recipient-target win by at least $0.50$ absolute relative to the clean recipient, while inactive-segment corruptions are not comparably damaging. \emph{Partial necessity} means the corruption effect has the same selective direction but does not reach that absolute-drop rule, or depends on an additional readout condition. Sim-CoT-GPT2 is therefore used as directional necessity evidence, not as a full sufficiency-and-necessity replication. Rows that do not pass the competence gate are competence boundaries; rows passing the gate with a different segment are carrier shifts rather than replications of the GPT-2 arithmetic mechanism.

\section{Experimental Setup}

We evaluate two six-step GPT-2-based local latent-reasoning checkpoints, CODI-GPT2 and Sim-CoT-GPT2. Important provenance disclosure: Sim-CoT-GPT2 is our Sim-CoT-style reproduction in the CODI-compatible code path, not the original released Sim-CoT checkpoint. Both checkpoints use the same latent-cache interface. Main localization uses $240$ held-out pairs for the CODI-GPT2 balls cell and two $60$-example seeds for the other GPT-2 template/checkpoint rows; matched-corruption tables aggregate $360$ examples over the three lexical variants; carrier-regime rows usually use $N=128$ held-out pairs per task. Appendix Table~\ref{tab:design-n} gives checkpoint provenance, sample sizes, seeds, and randomization details.

The main text reports point estimates because the carrier calls are separated by large oracle-versus-control gaps rather than by close threshold decisions. Appendix captions report row-level SEMs or bootstrap intervals where they are most relevant, and Table~\ref{tab:metric-audit} audits the decision thresholds and tie behavior used for the large-$N$ carrier map.

Both checkpoints are evaluated on the balls, books, and coins lexical variants. The main experiments use held-out source--recipient pairs, the fixed teacher-forced scoring rule from Section~\ref{sec:method}, and decoded free-generation checks for the key controls. The 1B/8B and task-breadth cells are used to map carrier regimes after competence gates.

The experimental battery follows one evidence order: sufficiency, component/source controls, matched corruption, decoded validation, and carrier-regime calibration. Appendix Table~\ref{tab:experiment-battery} gives the detailed battery by evidential role, while Appendix Table~\ref{tab:calibration-mini} classifies broader cells into latent-tail positives, carrier shifts, no-call boundaries, and reporting audits. Targeted repair is used only to satisfy the prerequisite that a causal diagnostic should be interpreted after the model can perform the relevant task mapping.

\begin{figure*}[!t]
\centering
\includegraphics[width=\textwidth]{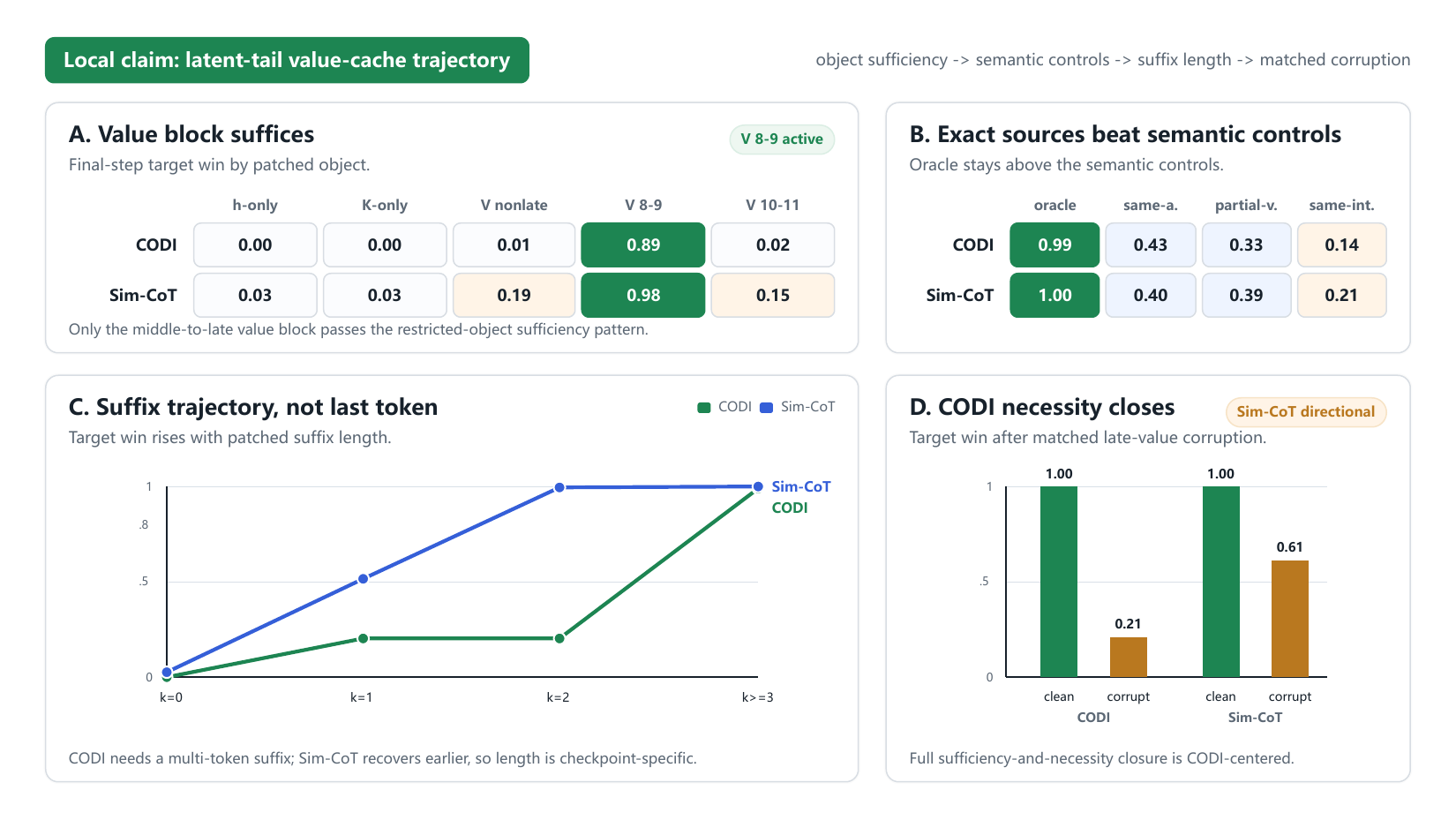}
\caption{Core arithmetic mechanism evidence. The panels read as a four-step evidence chain for the GPT-2 arithmetic row in Table~\ref{tab:headline-results}: the sufficient restricted object is the middle-to-late value block, semantic source controls fall below exact counterfactual sources, suffix-window curves show that the effect is not merely the final cache token, and matched corruption closes necessity only for CODI-GPT2. Unless otherwise marked, plotted values are final-step target-win rates. Numeric provenance is direct: layer values come from Table~\ref{tab:localization}; component controls from Tables~\ref{tab:component-sufficiency}, \ref{tab:vector-cache}, and~\ref{tab:simcot-vector-cache}; semantic controls from Table~\ref{tab:semantic}; suffix-window curves from Table~\ref{tab:suffix-window-main}; and corruption from Tables~\ref{tab:necessity} and~\ref{tab:simcot-necessity}.}
\label{fig:core-mechanism}
\end{figure*}

Each main GPT-2 diagnostic run uses a single GPU. The full evidence package is larger because it includes targeted repair, scale, and boundary checks, but the core intervention does not require a large distributed setup. We log resources per job rather than reporting a single normalized GPU-hour total across optional repair and calibration runs; the reproducibility appendix lists the run families and configs needed to regenerate the tables. The same intervention code runs on the public and matched LLaMA3-1B cells and on matched LLaMA3-8B repair/carrier-split cells; the larger checkpoints are used to test whether the causal carrier persists, shifts, or fails under modern model scale.

The full grid is a validation suite rather than the minimum practical protocol. A predeclared staged replay first tests oracle K/V over latent suffix, prompt prefix, and whole cache, then runs key-only and value-only checks only on the selected segment. On the 12 retained large-$N$ carrier cells, this replay recovers all 12 full-grid calls (5 latent-tail and 7 prompt-prefix) while reducing oracle segment/component localization from 9 to 5 configurations per cell, or 108 to 60 overall (44.4\%). All 16 audited carrier-threshold settings and competence floors $.70/.80/.90$ preserve 12/12 agreement. Configuration savings reduce wall time only when clean states are reused: naive staging changes runtime by at most 1.8\%, whereas single-process staging with cached clean states reduces wall time by 34.6--40.0\% on one matched 8B prompt-carrier cell and raises peak CUDA allocation from 16.5 to 17.3~GiB. This accounting does not treat competence, semantic-source, decoded, corruption, or optional layer/head checks as savings; those remain gate or finalist-only validation, and the timing result is not an end-to-end or frontier-scale claim.

\section{Results}

\begin{figure*}[!t]
\centering
\includegraphics[width=\textwidth]{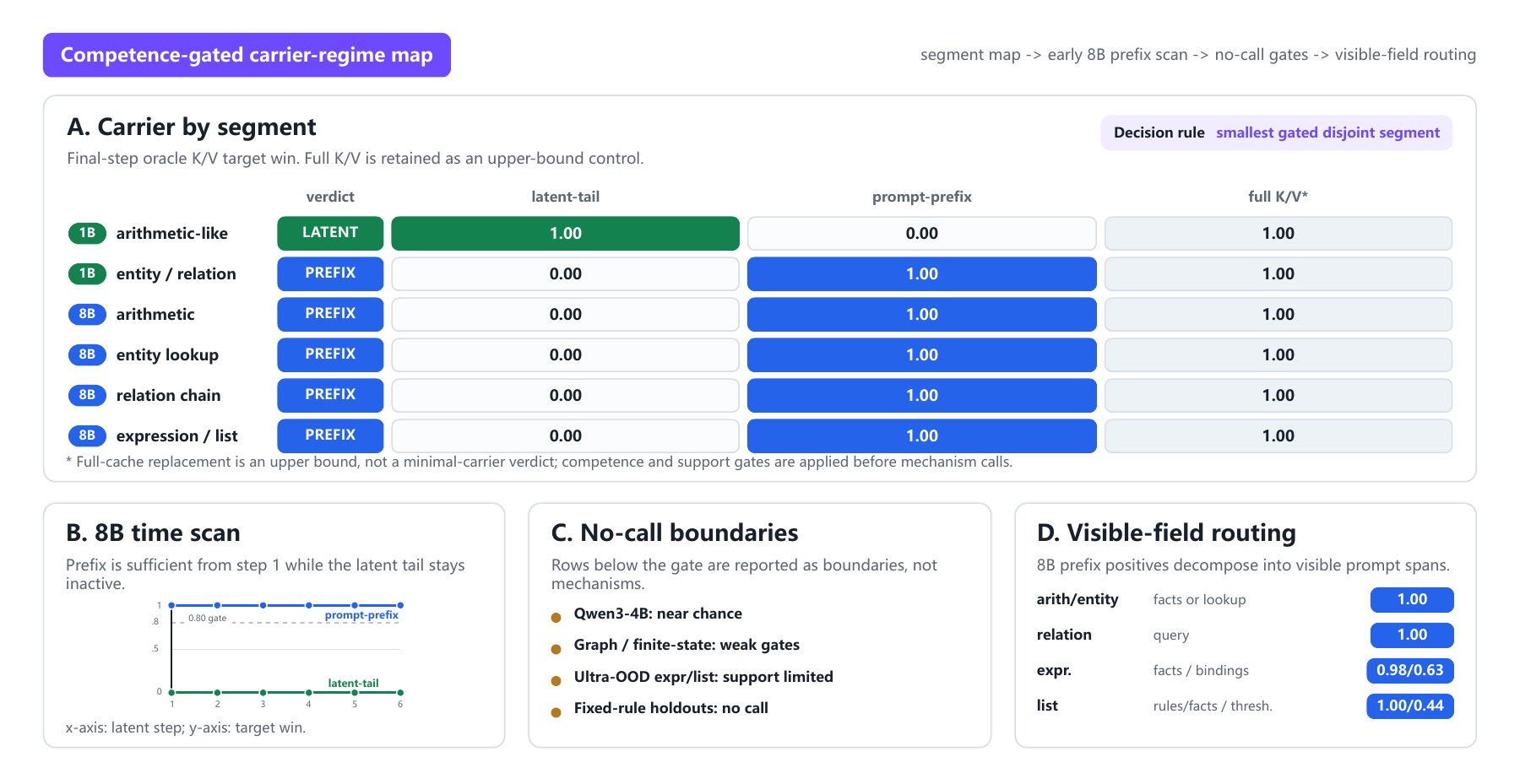}
\caption{Competence-gated carrier-regime map. The segment heatmap shows final-step target-win rates for latent-tail, prompt-prefix, and full-cache K/V; full-cache replacement is an upper-bound control, while the verdict follows the smallest gated disjoint segment. The 8B time scan shows prompt-prefix sufficiency from the first scanned latent step, the boundary panel lists no-call cells blocked by competence or support gates, and the field-routing inset summarizes the visible prompt spans behind the 8B prefix-shift rows.}
\label{fig:carrier-regime-main}
\end{figure*}

Table~\ref{tab:headline-results} is a claim dashboard: it states each verdict, points to the smallest evidence bundle, and the subsections below unpack the local mechanism, carrier-regime boundaries, and reporting audits. Figure~\ref{fig:core-mechanism} visualizes the GPT-2 arithmetic mechanism, Figure~\ref{fig:carrier-regime-main} visualizes the scale/task carrier map, and Appendix Table~\ref{tab:carrier-evidence-ledger} gives the full gate, sufficiency, corruption, decoded-behavior, and metric-audit ledger. The carrier-regime figure keeps the no-call rows separate from the visible-field routing panels so boundary cells are not read as mechanisms.

This organization is meant to make the main reviewer-relevant comparisons explicit: oracle interchange is compared against component, semantic-source, decoded, and corruption controls; cross-scale rows are interpreted only after competence gating; and shifted/no-call cells are reported as limits on the mechanism claim rather than as failed replications hidden in the appendix.

\subsection{Local arithmetic carrier}

Across the two GPT-2-scale arithmetic checkpoints, the active local carrier is a middle-to-late value-cache suffix trajectory. CODI-GPT2 reaches $0.875$--$0.908$ target win in the final-step layers 8--9 value block, while the final 10--11 block is $0.017$--$0.029$ and nonlate patches are $0.000$--$0.008$; Sim-CoT-GPT2 shows the same direction, with layers 8--9 at $0.958$--$0.992$ and 10--11 at $0.133$--$0.192$ (Appendix Tables~\ref{tab:localization}, \ref{tab:gpt2-block-scan}). Hidden-only, key-only, same-answer, partial-variable, cross-template partial, and decoded controls remain weaker, ruling out the current hidden vector, key-side routing, or a reusable answer/intermediate slot as the main explanation.

The trajectory evidence is strongest in CODI-GPT2: a one-token suffix reaches only $0.203$, while $k\geq3$ suffixes recover $0.984$--$1.000$ target win. Sim-CoT-GPT2 recovers earlier ($k=1$: $0.516$, $k=2$: $0.995$), so the shared claim is value-cache suffix transfer, not identical trajectory length. Matched corruption closes the sufficiency-and-necessity loop only for CODI-GPT2: late-value corruption drops recipient target win to $0.211$ with a $13.655$ log-probability drop, while Sim-CoT-GPT2 drops only to $0.606$ and is reported as directional necessity support (Appendix Tables~\ref{tab:necessity}, \ref{tab:simcot-necessity}). Decoded free-generation preserves the CODI separation under cache-only patching, while the enlarged-candidate stress test preserves the oracle-control margin instead of collapsing the four-answer setup (Appendix Tables~\ref{tab:freegen-cacheonly}, \ref{tab:freegen}; Figure~\ref{fig:candidate-stress}). Auxiliary noise, gradient, and head-Pareto checks agree with this narrower claim (Appendix Tables~\ref{tab:noise-prefix}, \ref{tab:value-gradient-readout}, \ref{tab:head-pareto}).

This establishes the local mechanism claim within its intended scope. The next subsection asks whether the same carrier survives once scale or task support changes.

\subsection{Carrier-regime boundaries}

The scale and task cells serve as scope calibration. Clean three-hop and public/matched 1B arithmetic-like cells preserve value/KV transfer, although their layer locus need not match GPT-2. Competent 8B arithmetic, entity, relation, and expression/list cells instead shift to prompt-prefix or full-cache K/V: Figure~\ref{fig:carrier-regime-main} shows prompt-prefix K/V at $1.000$ and latent-tail K/V at $0.000$ for the main 8B rows. The follow-ups make this a visible-field carrier diagnosis rather than a latent-tail accumulation account: prompt-prefix K/V is sufficient from the first scanned latent step, the no-latent probe shows that the prompt state is already sufficient before rollout, and post-prefix ablations show that task-relevant visible fields remain readout-active after latent encoding (Appendix Tables~\ref{tab:main-8b-time-prefix}, \ref{tab:8b-no-latent}, \ref{tab:8b-prefix-readout-ablation}). Full-cache positives are retained in the appendix ledger as upper-bound checks. The remaining ambiguity is timing: the answer may already be completed before latent rollout, or later readout may consult prompt fields that were already encoded.

The field split makes the 8B carrier shift less opaque: Table~\ref{tab:field-routing-main} identifies the visible spans behind the shift instead of treating the 8B result as an opaque full-cache positive.
\begin{table}[H]
\centering
\footnotesize
\setlength{\tabcolsep}{2pt}
\renewcommand{\arraystretch}{0.90}
\begin{tabularx}{\linewidth}{@{}L{0.27\linewidth}c c Y@{}}
\toprule
\textbf{8B cell} & \textbf{Lat.} & \textbf{Pref.} & \textbf{Best visible field(s)} \\
\midrule
2-hop arithmetic & .00 & 1.00 & facts / initial clause: 1.00 \\
Entity lookup & .00 & 1.00 & lookup entry / facts: 1.00 \\
Relation chain & .00 & 1.00 & query: 1.00; tables: .00 \\
Expression tree & .00 & .99 & facts: .98; bindings: .63 \\
List filter/sum & .00 & 1.00 & rules/facts: 1.00; threshold: .44 \\
\bottomrule
\end{tabularx}
\caption{Main-text slice of 8B field routing. Lat. and Pref. are oracle target-win rates for latent-tail and prompt-prefix K/V replacement. Visible fields are the strongest prompt-span patches; source-control rows, OOD variants, and post-prefix readout ablations are in Appendix Tables~\ref{tab:field-routing-hardening} and~\ref{tab:8b-prefix-readout-ablation}.}
\label{tab:field-routing-main}
\end{table}

The relation-chain task provides the clearest non-arithmetic exact-control instance. Executable person$\rightarrow$badge$\rightarrow$room relations provide verified recipient, counterfactual, and control labels. At both 1B and 8B scale, latent/prompt/whole-cache K/V target win is $0/1/1$; at 8B, decoded transfer is $0/1$ for latent/prompt patching and recipient preservation after matched corruption is $1/0$. Across three semantically equivalent renderings, every competence-passing cell retains the same prompt-prefix carrier call (Appendix Table~\ref{tab:relation-surface}). Thus sufficiency, decoded transfer, corruption, and surface variation all select visible prompt/source binding rather than the arithmetic late-value mechanism.

No-call cells are withheld rather than folded into the mechanism claim. Qwen3-4B probes, graph path, broader finite-state cells, ultra-OOD expression/list variants, fixed-rule holdouts, and high-random-baseline wide-OOD rows are boundary or secondary evidence (Appendix Tables~\ref{tab:failure-mini}, \ref{tab:repair-boundary}, \ref{tab:fixedrule-support-boundary}, \ref{tab:llama1b-stress}).

\paragraph{Variable-stop pilot.}
We froze an evaluator-defined stopping rule before evaluating a held-out $N=48$ slice; source and recipient stops differed for 27/48 pairs. Reverse-aligned latent K/V+$h$ transfer succeeded on 47/48 pairs overall and 26/27 unequal-stop pairs, and $k=2$ was the shortest tested window supported in both equal- and unequal-stop groups. Prompt K/V alone succeeded on 0/48, while latent K/V alone and $h$ alone reached 27/48 and 14/48. This supports a joint reverse-aligned latent K/V+$h$ carrier in this pilot, not value-specific necessity or a stopping-policy-invariant carrier (Appendix Table~\ref{tab:adaptive-stop}).

These boundary calls motivate the final checks: one asks whether the carrier is stable across training path, and the other asks whether the carrier calls depend on the stated reporting rules.

\subsection{Path sensitivity and reporting audits}

Training-path probes make the generalization limit explicit. Range-augmented $T=16$ gives two competent seeds with different carrier basins, and three newly trained CODI-style seeds have high clean accuracy but weak late-8--9 value sufficiency ($0.049$--$0.138$) and weak corruption drops ($0.026$--$0.089$). Their latent/prompt/whole-cache K/V target-win rates are $.370/.203/.870$, $.208/.427/.865$, and $.427/.141/.906$: whole-cache transfer is present, but neither disjoint segment reaches the $.80$ carrier gate, so all three receive no localized-carrier call. This excludes a simple shift to a strong prompt-prefix carrier and further bounds the headline mechanism to the main CODI-GPT2 checkpoint (Appendix Tables~\ref{tab:training-horizon-boundary}, \ref{tab:codi-multiseed-bootstrap}).

The reporting audits do not change the headline carrier split. On the retained large-$N$ 1B/8B rows, sweeping the competence gate over $.70/.80/.90$ and crossing sufficiency floors $.75$--$.90$ with competitor ceilings $.10$--$.30$ leaves the same $5$ latent-tail and $7$ prompt-prefix calls; exact top-score ties occur in only $.063\%$ of retained cache-interchange score rows (Appendix Table~\ref{tab:metric-audit}). These audits do not make the mechanism universal, but they rule out the most direct threshold and tie-breaking explanations for the displayed carrier map.

\section{Discussion}

Latent-CoT causal analysis should identify the internal object being tested, not only the latent step. In the GPT-2 arithmetic checkpoints studied here, the transferable object is not the final hidden state alone, key-side routing alone, or a reusable answer/intermediate slot; the best operational description is the context-bound value-cache trajectory defined in Section~\ref{sec:method}. This is a controlled sufficiency-and-control claim, not a decoded symbolic trace or a named single-head circuit. Component splits should be read with the same intervention semantics: value-only success means source values are sufficient under recipient routing and downstream readout, not that natural inference literally uses isolated values while keys are inert.

The most important negative result is the 8B prompt-prefix regime. Section 4 shows that the carrier shifts to visible prompt fields, and the no-latent truncation probe shows that the prompt-prefix state is already sufficient before latent rollout. The post-prefix ablation follow-up adds the missing direction: after latent encoding, zeroing prompt-prefix K/V or task-relevant visible spans collapses answer readout, while zeroing the latent suffix does not. This supports visible-field readout as part of the 8B mechanism, but it does not exclude early completion because the pre-rollout prompt state is already sufficient. The role left for latent iterations in these cells is therefore narrower: they are not where the counterfactual answer is newly stored in tail K/V, though they may still implement readout or control operations over an already informative prompt-prefix cache.

The right generalization target is therefore a competence-gated carrier-regime map, not a universal carrier. \method{} is complementary to Coconut, CODI, and Sim-CoT training papers \citep{hao2024training,shen2025codi,wei2025simcot}: it asks which transformer object carries a known counterfactual computation after a model is trained. The present evidence supports a focused GPT-2 arithmetic mechanism and a broader diagnostic framework; it does not predict the carrier of a new task or model before interventions, and the seed/path audits rule out using a single latent-tail map as a deployment guarantee.

Looking forward, a controlled GSM8K extension would begin from a predeclared subset encoded as executable solution graphs. Perturbing one numeric leaf and re-executing its downstream nodes would provide verified recipient and counterfactual answers. Graph edits would be crossed independently with paraphrase, discourse-order, and distractor variants, while source--recipient pairing remains within each rendering. Competence and template-coverage gates would precede localization, and semantic-source, decoded-transfer, and matched-corruption controls would be retained. This is an evaluation pathway, not evidence of GSM8K robustness in the present paper.

For loop transformers or recurrent models, \method{} could map carriers by iteration or reverse distance from the realized stop rather than fixing one carrier throughout training. A training-only extension could optimize $\mathcal{L}=\mathcal{L}_{\mathrm{clean}}+\lambda\mathcal{L}_{\mathrm{interchange}}$, where verified source interventions target their corresponding counterfactual labels and inactive patches preserve the recipient label. The carrier would be periodically remapped, while unseen depths, loop counts, seeds, and matched-corruption necessity remain evaluation criteria.

\section*{Limitations}

The experiments are controlled by design. All cells focus on synthetic tasks with exact counterfactual semantics, because \method{} needs known source--recipient answers and partial-variable controls. Balls, books, and coins are lexical surface variants of one two-hop arithmetic diagnostic, not a broad benchmark suite. The expanded cells reduce but do not remove this task-breadth limitation: clean three-hop arithmetic gives positive all-layer transfer, while conditional arithmetic, entity lookup, expression/list repair, relation-chain repair, finite-state controls, graph path, and fixed-rule holdouts expose carrier shifts or competence boundaries. These cells prevent overgeneralization; they do not establish broad task generality or latent-tail reasoning outside the arithmetic regime. Mechanisms on synthetic two-hop arithmetic may differ from mechanisms used on natural GSM8K-like multi-step word problems, whose language variability, distractors, and solution strategies do not come with exact source--recipient counterfactuals or partial-variable controls.

The variable-stop pilot freezes the clean stopping decisions and decodes at the recipient stop; it does not rerun the stopping controller after intervention. It covers one synthetic checkpoint with realized stops 3--5 and provides no matched-corruption necessity evidence. Its joint K/V+$h$ call is therefore specific to the tested stopping policy and should not be read as learned, native, or open-ended halting support.

The OOD numeric-range check reduces the risk that the localization is a small-integer artifact, but its random-source baseline is high, so it is secondary. Public and matched LLaMA3-1B diagnostics support all-layer value-cache transfer, but the strongest restricted region shifts relative to GPT-2. The matched LLaMA3-8B split supports full-cache/prompt-prefix transfer, not latent-tail transfer. Sim-CoT also depends more on the source hidden state at decode time than CODI-GPT2, so cache-only sufficiency is not universal across checkpoints.

The strongest sufficiency-plus-necessity arithmetic claim is also not a recipe-level replication claim. Sim-CoT-GPT2 supplies strong sufficiency and decoded-control support but only directional matched-corruption damage, and newly trained CODI-style seeds with high clean accuracy fail to reproduce the original checkpoint's late-8--9 value carrier. We therefore treat the original CODI-GPT2 closure as evidence for a realized mechanism in that checkpoint, not as a stable property of the CODI training recipe.

The Qwen3-4B row is a calibration boundary rather than mechanism evidence: it differs from the LLaMA-family rows in architecture, tokenizer, checkpoint provenance, and repair recipe, so the apparent 1B-pass, Qwen3-4B-fail, 8B-pass-with-shift pattern should not be read as a scale law. This is not an explanation of the 4B failure; it is a confound audit. Resolving the nonmonotonic-looking pattern would require matched architecture families, tokenizers, training recipes, repair support, and competence gates across sizes.

Fully resolving scale generalizability would require the same competence-gated diagnostic battery across several architectures and sizes, with architecture-relative layer partitions rather than a fixed GPT-2 notion of ``late.'' Finally, although we validate the main controls under greedy decoding and a limited temperature-sampling diagnostic, broader unconstrained decoding and downstream task behavior should be tested before making broad claims about deployment behavior.

\section*{Ethics Statement}

This work analyzes small reasoning models on synthetic arithmetic data and does not use personal or sensitive data. The main risk is overclaiming interpretability: causal localization in controlled settings should not be presented as a general guarantee of faithful reasoning in deployed models. We therefore restrict our claims to the tested checkpoints, templates, and intervention definitions.

\bibliography{custom}

\appendix

\section{Evidence Appendix}

\subsection{How to Read the Appendix}
\label{app:appendix-guide}

The appendix is organized by evidential role rather than by execution order. Table~\ref{tab:appendix-map} gives the reading map; the main entry points are the expanded schematic, Table~\ref{tab:concern-evidence-map}, the carrier ledger, the 8B field-routing table, the metric audit, and the carrier split. No-call is a reporting category, not a single failure mode: competence, support, source-control, and high-baseline OOD rows are all withheld for different reasons.

Unless otherwise stated, entries written as \twostep{x}{y} report latent steps 5/6. Target win (TW) is the closed-set teacher-forced winner rate defined in Section~\ref{sec:method}. The appendix tables are not an additional benchmark leaderboard: repaired or augmented cells calibrate carrier regimes after competence gates, while failed-gate cells define boundaries.

\begin{table*}[t]
\centering
\appendixtablefont
\setlength{\tabcolsep}{4pt}
\begin{tabularx}{\textwidth}{@{}L{0.21\textwidth}L{0.33\textwidth}Y@{}}
\toprule
\textbf{Appendix block} & \textbf{Tables / figures} & \textbf{Purpose} \\
\midrule
Method/setup & Main-text Figure~\ref{fig:scit}; Appendix Figure~\ref{fig:scit-expanded}; Tables~\ref{tab:intervention-defs}--\ref{tab:design-n} & Defines the intervention grid, prompt templates, scoring/battery, and sample sizes moved out of the main text. \\
Carrier-regime audit & Tables~\ref{tab:calibration-mini}--\ref{tab:evidence-map} & Gives the carrier-status index, 8B time scan, no-latent truncation and post-prefix readout probes, no-call rows, concern-to-evidence map, ledger, field routing, threshold/tie audit, and method-positioning map. \\
Core arithmetic & Tables~\ref{tab:localization}--\ref{tab:necessity}; Figure~\ref{fig:candidate-stress} & Supplies the GPT-2 localization, semantic-source controls, matched corruption, and candidate-set stress check behind the local mechanism claim. \\
Scope/boundaries & Tables~\ref{tab:benchmark-scope}--\ref{tab:fixedrule-support-boundary} & Separates task-breadth calibration, repaired cells, support boundaries, and base-model sanity checks. \\
Trajectory/auxiliary checks & Tables~\ref{tab:trajectory-content}--\ref{tab:codi-multiseed-bootstrap} & Collects trajectory-length, head, geometry, noise, gradient, ablation, horizon, and seed/path diagnostics. \\
Carrier-scale/component details & Tables~\ref{tab:carrier-split}--\ref{tab:simcot-necessity} & Expands segment splits, GPT-2 block scans, OOD/1B checks, vector/cache controls, cross-template controls, decoded validation, and Sim-CoT corruption. \\
\bottomrule
\end{tabularx}
\caption{Appendix table map. Each row names the appendix block and the tables or figures that support it.}
\label{tab:appendix-map}
\end{table*}

\subsection{Method, Setup, and Scoring Details}
\label{app:method-setup}

\subsubsection{Intervention and Prompt Definitions}
\begin{figure*}[t]
\centering
\includegraphics[width=0.92\textwidth]{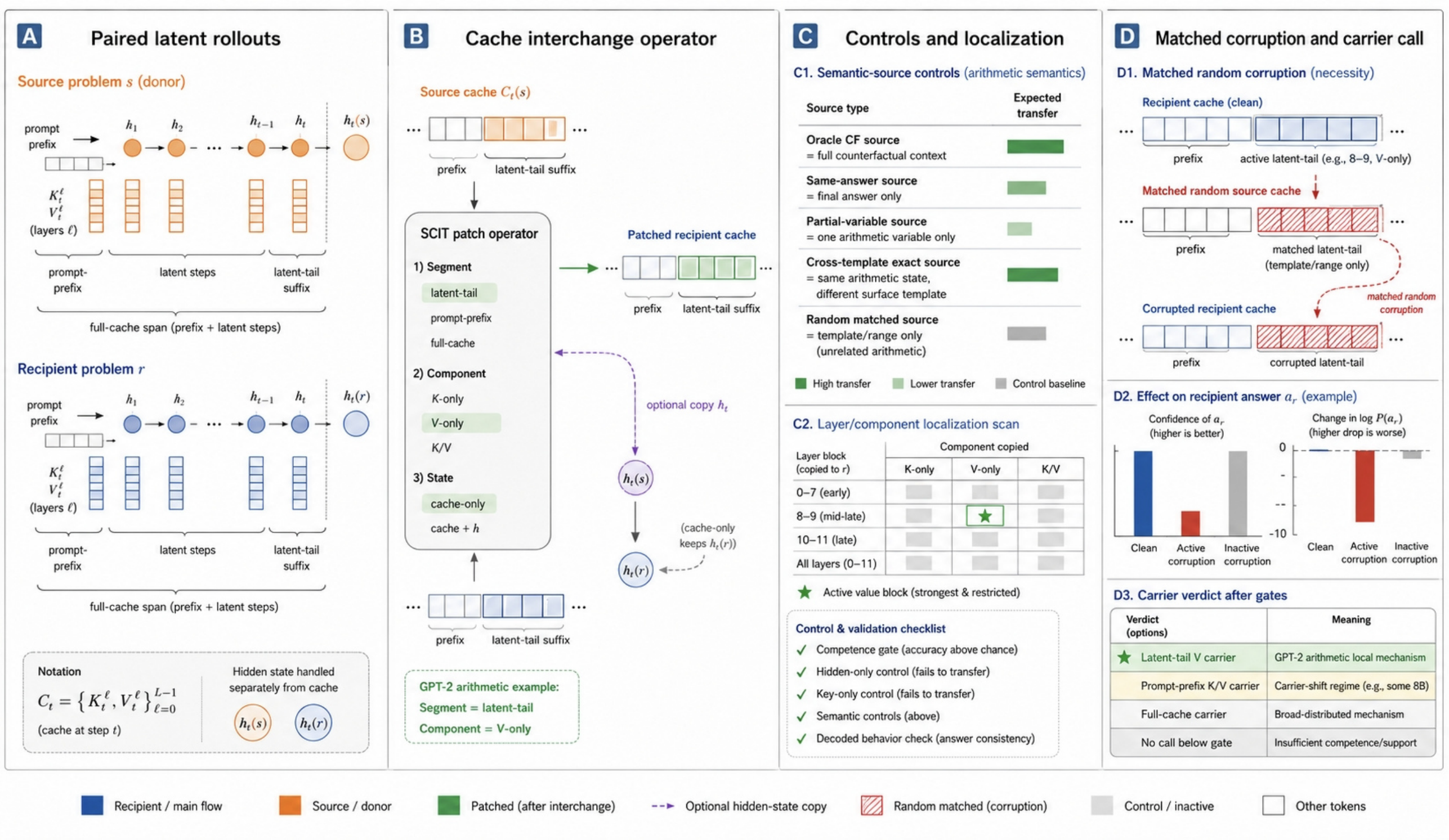}
\caption{Expanded \method{} schematic. The main-text Figure~\ref{fig:scit} gives the compact carrier-test protocol; this appendix figure expands the same contract into cache-slice interchange, intervention-option choices, semantic-source controls, matched random corruption, and scoring/readout details.}
\label{fig:scit-expanded}
\end{figure*}

Figure~\ref{fig:scit-expanded} and Tables~\ref{tab:intervention-defs}--\ref{tab:design-n} contain details compressed out of the main Method and Experimental Setup sections: intervention definitions, prompt templates, the experimental battery, and sample-size/randomization choices.

\begin{table*}[t]
\centering
\appendixtablefont
\setlength{\tabcolsep}{3pt}
\begin{tabularx}{\textwidth}{@{}L{0.15\textwidth}L{0.23\textwidth}L{0.20\textwidth}L{0.15\textwidth}Y@{}}
\toprule
\textbf{Name} & \textbf{Cache segment} & \textbf{Component operation} & \textbf{Current state} & \textbf{Interpretation} \\
\midrule
Hidden-only & Recipient cache unchanged & No K/V replacement & Replace $h_t(r)$ with $h_t(s)$ & Tests whether the current hidden vector alone carries the counterfactual. \\
Cache-only & Selected segment from source & Replace K and V jointly unless restricted below & Keep $h_t(r)$ & Tests whether the cache trajectory is sufficient without copying the source hidden state. \\
Cache+h & Selected segment from source & Replace K and V jointly unless restricted below & Replace $h_t(r)$ with $h_t(s)$ & Strong closed-loop interchange; interpreted only with hidden-only and cache-only controls. \\
Key-only & Selected segment from source & Replace K; keep recipient V & As specified by cache-only/cache+h & Surgical routing test; an artificial state used to test key-only sufficiency. \\
Value-only & Selected segment from source & Replace V; keep recipient K & As specified by cache-only/cache+h & Surgical content test; success means source value content is usable under recipient keys. \\
Full-cache & Prompt prefix plus latent tail & Replace all available K/V cache entries & As specified by cache-only/cache+h & Upper-bound interchange check and comparison point for segment-local claims. \\
\bottomrule
\end{tabularx}
\caption{Intervention definitions. The rows are not all mutually exclusive experimental families: key-only, value-only, and K/V are component choices that can be crossed with cache-only or cache+h state handling. Segment choices such as latent-tail, prompt-prefix, and whole-cache determine where the component operation is applied. Cache-only rows copy K/V entries at the patch point but keep the recipient's current hidden state $h_t$; hidden-only rows copy $h_t$ but leave all K/V entries unchanged.}
\label{tab:intervention-defs}
\end{table*}

\begin{table*}[t]
\centering
\appendixtablefont
\setlength{\tabcolsep}{3pt}
\begin{tabularx}{\textwidth}{@{}L{0.12\textwidth}Y@{}}
\toprule
\textbf{Family} & \textbf{Prompt template} \\
\midrule
balls & \texttt{Question: Alice has \{A\} red balls and \{B\} blue balls. She gives away \{C\} red balls. How many balls does she have now? Answer:} \\
books & \texttt{Question: Alice has \{A\} science books and \{B\} history books. She lends out \{C\} science books. How many books does she have now? Answer:} \\
coins & \texttt{Question: Alice has \{A\} silver coins and \{B\} gold coins. She gives away \{C\} silver coins. How many coins does she have now? Answer:} \\
\bottomrule
\end{tabularx}
\caption{Arithmetic prompt templates. The implementation places \texttt{Answer:} on a new line; the table uses one line only for compactness. If $P$ is the post-transfer count, the variables satisfy $A=P+C$ and the answer is $P+B$. Books and coins preserve the same computation while changing the surface noun family and transfer verb.}
\label{tab:prompt-templates}
\end{table*}

\subsubsection{Scoring and Semantic Source Construction}
For arithmetic examples, write the transferable count as $A$, the transfer-away count as $C$, the post-transfer intermediate as $P=A-C$, and the held-fixed count as $B$, so the answer is $P+B$. The oracle source combines donor $P_s$ with recipient $B_r$; e.g., $B_r=8$ and $P_s=11$ gives oracle answer $19$. Same-answer controls keep only the final answer, same-intermediate controls keep $P_s$ while changing $B$, partial-variable controls keep $(P_s,B_r)$ while changing the surface decomposition $(A,C)$, and cross-template exact/partial controls move the same distinction across books or coins.

Localization experiments score fixed candidate answers with summed teacher-forced log-probabilities from the patched state. The default candidate set is recipient, donor, oracle-counterfactual, and random-control answers; semantic controls add a distinct source answer when needed. The continuous margin used in high-baseline or enlarged-candidate checks is the same $\Delta_{\mathrm{cf-rec}}$ defined in Section~\ref{sec:method}. For matched random corruption, the expected target is the recipient answer $a_r$; random sources are matched by template family, numeric range, latent step, and cache segment shape, with recipient, donor, and oracle answers excluded.

\begin{table*}[t]
\centering
\scriptsize
\setlength{\tabcolsep}{2pt}
\begin{tabularx}{\textwidth}{@{}L{0.13\textwidth}L{0.22\textwidth}L{0.22\textwidth}Y@{}}
\toprule
\textbf{Family} & \textbf{Cells} & \textbf{Intervention/readout} & \textbf{Evidence role} \\
\midrule
Core localization & CODI-GPT2 and Sim-CoT-GPT2 on balls/books/coins & Latent-tail cache+h over layer groups; cache-only, hidden-only, key-only, value-only splits & Identifies the arithmetic carrier as a value-cache trajectory rather than the current hidden vector or key routing. \\
Semantic controls & Exact, same-answer, partial-variable, cross-template, random, greedy, sampled, enlarged-candidate checks & Same patched state scored by target win, margin, and parsed free generation & Tests whether transfer is full-context counterfactual behavior rather than answer-slot reuse or a closed-set artifact. \\
Necessity & Matched random corruption plus zero/mean ablations & Recipient-answer preservation and log-probability drop after damaging the same region & Separates sufficiency from requirement and checks whether the localized region is globally fragile. \\
Carrier-regime calibration & 1B/8B arithmetic, three-hop, entity, relation, expression/list, state, graph, rule cells & Latent tail, prompt prefix, whole cache, component, decoding, support, and OOD gates & Maps whether the carrier persists, shifts, or becomes uninterpretable; these cells are not treated as broad benchmark wins. \\
8B hardening & Field-level 8B arithmetic, entity, relation, expression/list probes with source controls & Field K/V patches over lookup, query, fact, rule, operand, threshold spans & Decomposes prefix or full-cache positives into visible-field carrier evidence rather than leaving 8B as an opaque stress test. \\
Reporting audits & Threshold sweeps, tie audit, head controls, seed/path probes & Fixed decision grid, tie counts, ranked versus random head patches, checkpoint scans & Checks that carrier calls do not depend on arbitrary thresholds, tie order, or a single training path. \\
\bottomrule
\end{tabularx}
\caption{Experimental battery by evidential role. The main mechanism claim uses the core localization, semantic controls, decoded validation, and matched corruption rows. Broader 1B/8B and task-breadth rows are competence-gated calibration.}
\label{tab:experiment-battery}
\end{table*}

The checkpoint and sample-size details behind Table~\ref{tab:design-n} are collected here. CODI-GPT2 is the CODI-style local adapter loaded from \texttt{models/CODI-gpt2}. Sim-CoT-GPT2 is not the original released Sim-CoT checkpoint; it is our GPT-2 reproduction in the same CODI-compatible code path, using the Sim-CoT decoder option and loaded from \texttt{models/sim-cot-gpt2-codi}.

Both checkpoints use the standard GPT-2 backbone family, LoRA adapters with rank $128$, $\alpha=32$, dropout $0.1$, a 768-dimensional latent projection, and six latent iterations at evaluation time. The configs \texttt{configs/scit/codi\_gpt2\_scit.yaml} and \texttt{configs/scit/simcot\_codi\_gpt2\_scit.yaml} specify the tokenizer, latent-step count, projection, LoRA settings, prompt template, and teacher-forced scoring template. The local adapters were trained from GSM8K/GSM8K-Aug style latent-reasoning supervision, so we report them as local reproductions rather than off-the-shelf CODI or Sim-CoT artifacts.

\begin{table*}[t]
\centering
\appendixtablefont
\setlength{\tabcolsep}{3pt}
\begin{tabularx}{\textwidth}{@{}L{0.18\textwidth}L{0.22\textwidth}L{0.18\textwidth}Y@{}}
\toprule
\textbf{Diagnostic} & \textbf{Main cells} & \textbf{$N$ / seeds} & \textbf{Randomization and readout} \\
\midrule
GPT-2 localization & CODI-GPT2 and Sim-CoT-GPT2, balls/books/coins & CODI balls: 240 examples per cell; other rows: two 60-example seeds & Held-out source--recipient pairs; target win over recipient, donor, oracle, and random labels; steps 5/6 reported. \\
Semantic controls & Same models/templates & Same as localization & Same-answer, same-intermediate, partial-variable, cross-template, and random sources; fifth label added when needed. \\
Matched corruption & CODI-GPT2 main; Sim-CoT-GPT2 replication & 360 examples aggregated over balls/books/coins per table & Same-template and same-range random cache sources; recipient-answer preservation and log-probability drop. \\
Decoded validation & Greedy and sampled free generation & Greedy: 60 examples per template; sampling: five samples per example & Parsed free-generation answer compared with counterfactual, recipient, and out-of-set labels. \\
Carrier-regime calibration & 1B/8B arithmetic, entity, relation, expression/list, state cells & Usually $N=128$ held-out pairs per carrier-map task; field rows caption exact $N$ & Competence-gated latent-tail, prompt-prefix, and whole-cache segment splits with source controls where available. \\
\bottomrule
\end{tabularx}
\caption{Experimental design details. Appendix captions give row-specific SEMs and exceptions; this table records the core $N$, seed, and randomization choices.}
\label{tab:design-n}
\end{table*}

\FloatBarrier

\subsection{Carrier Map and Reporting Audits}
\label{app:audit-index}

\subsubsection{Carrier-Status Mini Tables}
Tables~\ref{tab:calibration-mini}--\ref{tab:failure-mini} summarize carrier-regime calls, the time-resolved 8B prompt-prefix follow-up, the no-latent truncation probe, post-prefix readout ablations, and no-call boundary rows before the appendix turns to the full audit ledger.

\begin{table}[t]
\centering
\appendixtablefont
\setlength{\tabcolsep}{3pt}
\begin{tabularx}{\linewidth}{@{}L{0.28\linewidth}L{0.30\linewidth}Y@{}}
\toprule
\textbf{Call type} & \textbf{Cells} & \textbf{Interpretation} \\
\midrule
Latent-tail positive & GPT-2 arithmetic; clean three-hop; 1B arithmetic-like & Arithmetic carrier persists, but layer locus need not be scale-invariant. \\
Carrier shift & 8B arithmetic, entity, relation; repaired expression/list; repaired relation-chain; state controls & Competent cells route through prompt-prefix or full-cache K/V; no-latent and post-prefix ablations show the prompt-prefix state is sufficient before rollout and visible fields remain readout-active afterward. \\
No-call boundary & Qwen3-4B, graph path, finite-state, ultra-OOD expression, list controls, fixed-rule holdouts & Competence, support, or source-control gates block mechanism interpretation. \\
Reporting audit & Threshold sweep, tie audit, head Pareto, seed/path probes & Checks that calls are not artifacts of cutoffs, tie order, or one training path. \\
\bottomrule
\end{tabularx}
\caption{Calibration-suite status. These cells calibrate carrier regimes; they are not treated as broad benchmark-generalization wins.}
\label{tab:calibration-mini}
\end{table}

\begin{table*}[t]
\centering
\scriptsize
\setlength{\tabcolsep}{2pt}
\begin{tabularx}{\textwidth}{@{}L{0.15\textwidth}cL{0.13\textwidth}L{0.14\textwidth}L{0.14\textwidth}L{0.12\textwidth}Y@{}}
\toprule
\textbf{8B follow-up cell} & \textbf{$N$} & \textbf{Latent-tail TW} & \textbf{Prompt-prefix TW} & \textbf{Prompt $\Delta_{\mathrm{cf-rec}}$} & \textbf{Top-8 attn. Lat./Pref.} & \textbf{Interpretation} \\
\midrule
Two-hop arithmetic & 48 & 0.000--0.000 & 1.000--1.000 & 26.2--26.2 & .013 / .960 & Prefix from step 1. \\
Entity lookup & 48 & 0.000--0.000 & 1.000--1.000 & 31.8--31.8 & .001 / .876 & Prefix lookup binding. \\
Expression tree & 48 & 0.000--0.000 & 1.000--1.000 & 35.6--35.6 & .004 / .987 & Prefix operand/fact binding. \\
Relation chain & 48 & 0.000--0.000 & 1.000--1.000 & 41.9--41.9 & .031 / .962 & Prefix query binding. \\
\bottomrule
\end{tabularx}
\caption{Time-resolved 8B prompt-prefix follow-up. Each row scans latent steps 1--6 with oracle cache-only K/V patches over the latent suffix and prompt prefix. Target-win ranges are the minimum and maximum over the six scanned steps: prompt-prefix K/V is sufficient from the first scanned latent step, while latent-suffix K/V is inactive throughout. The attention column reports final-step answer-readout mass for top-8 heads selected by an independent prompt-prefix K/V head scan, averaged across recipient and oracle-counterfactual states.}
\label{tab:main-8b-time-prefix}
\end{table*}

\begin{table*}[t]
\centering
\scriptsize
\setlength{\tabcolsep}{2pt}
\begin{tabularx}{\textwidth}{@{}L{0.17\textwidth}cL{0.13\textwidth}L{0.13\textwidth}L{0.13\textwidth}L{0.13\textwidth}Y@{}}
\toprule
\textbf{8B truncation cell} & \textbf{$N$} & \textbf{Direct pre-recipient TW} & \textbf{Direct pre-oracle TW} & \textbf{Latent-tail TW} & \textbf{Full-cache TW} & \textbf{Interpretation} \\
\midrule
Two-hop arithmetic & 32 & 1.000 & 1.000 & 0.000 & 1.000 & Prefix state is sufficient before rollout. \\
Expression tree & 32 & 1.000 & 0.969 & 0.000 & 0.969 & One oracle miss; same qualitative prefix regime. \\
List filter/sum & 32 & 1.000 & 1.000 & 0.000 & 1.000 & Prefix state dominates latent tail. \\
Relation chain & 32 & 1.000 & 1.000 & 0.000 & 1.000 & Query/source binding is pre-rollout sufficient. \\
\bottomrule
\end{tabularx}
\caption{8B no-latent truncation probe. Direct-pre rows score candidate answers from the state returned by \texttt{forward\_until\_step} before completing the latent rollout; latent-tail and full-cache columns are step-1 oracle K/V intervention controls from the same held-out run. The prompt-prefix state already selects the recipient answer under the recipient prompt and the counterfactual answer under the oracle prompt, while latent-tail patches remain inactive. This rules out a necessary latent-tail accumulation account; by itself, the truncation probe does not separate early completion from later visible-field readout.}
\label{tab:8b-no-latent}
\end{table*}

\begin{table*}[t]
\centering
\scriptsize
\setlength{\tabcolsep}{2pt}
\begin{tabularx}{\textwidth}{@{}L{0.15\textwidth}cL{0.13\textwidth}L{0.13\textwidth}L{0.15\textwidth}L{0.15\textwidth}Y@{}}
\toprule
\textbf{8B readout cell} & \textbf{$N$} & \textbf{Latent-suffix zero TW} & \textbf{Prompt-prefix zero TW} & \textbf{Best field zero TW} & \textbf{Unaffected field} & \textbf{Interpretation} \\
\midrule
Two-hop arithmetic & 24/12 & 1.000 & .417 / .417 & initial/facts: .417--.500 & query: 1.000 & Numeric facts remain required at readout. \\
Expression tree & 24/12 & 1.000 & .458 / .458 & rules: .250; inputs: .417 & query: 1.000 & Rules and bindings, not latent suffix, support answer readout. \\
List filter/sum & 24/12 & 1.000 & .125 / .125 & values: .583; facts: .750 & query/rules: .917--1.000 & Value bindings dominate the visible-field dependency. \\
Relation chain & 24/12 & 1.000 & .250 / .250 & room/query: .167--.250 & person table: 1.000 & Readout depends on the queried relation fields. \\
\bottomrule
\end{tabularx}
\caption{8B post-prefix readout ablations. The first two target-win columns use source-free zero ablation over the recipient latent suffix or prompt prefix at latent step 6, shown as no-continue / continue values for K/V ablation. The field column uses a smaller localization probe ($N=12$) that zeroes visible prompt spans at direct-pre, no-continue, and post-latent-final stages; the displayed ranges summarize the damaging K/V field spans across those stages. These are harsher ablations than matched corruption, so we use them as readout-sensitivity evidence rather than counterfactual-transfer evidence. They show that latent suffix entries are not required, while prompt-prefix and task-relevant visible fields remain required after latent encoding.}
\label{tab:8b-prefix-readout-ablation}
\end{table*}

\begin{table}[t]
\centering
\appendixtablefont
\setlength{\tabcolsep}{3pt}
\begin{tabularx}{\linewidth}{@{}L{0.32\linewidth}L{0.29\linewidth}Y@{}}
\toprule
\textbf{Boundary cell} & \textbf{Gate failure} & \textbf{Interpretation} \\
\midrule
Qwen3-4B probes & Closed-set competence near chance & No interpolation claim between 1B and 8B. \\
Graph path & Weak decoded competence & Path-following training support is insufficient. \\
Broader finite-state & Short/long sequence gates fail & State tracking does not generalize from narrow supports. \\
Expression ultra-OOD & Free-generation and source-control gates fail & Repaired carrier is support-limited. \\
List hard cells & Source controls remain high & Prompt-prefix copying cannot be separated cleanly. \\
Fixed-rule holdouts & Leave-one/class holdouts fail & Matched-support fitting, not robust rule execution. \\
\bottomrule
\end{tabularx}
\caption{No-call boundary cells. Each row fails a prerequisite gate, so \method{} withholds a latent-tail mechanism call.}
\label{tab:failure-mini}
\end{table}

\FloatBarrier

\subsubsection{Concern-to-Evidence Map}
Table~\ref{tab:concern-evidence-map} maps likely weaknesses to compact supporting diagnostics, so the expanded appendix is easier to scan.

\begin{table*}[t]
\centering
\appendixtablefont
\begin{tabularx}{\textwidth}{@{}L{0.26\textwidth}L{0.30\textwidth}Y@{}}
\toprule
\textbf{Weakness} & \textbf{Where to look} & \textbf{What the evidence supports} \\
\midrule
The main task is too narrow & Tables~\ref{tab:benchmark-scope}, \ref{tab:generalization-roadmap}, \ref{tab:scit-bench-v2}, \ref{tab:repair-boundary} & The paper is a causal diagnostic, not a broad reasoning benchmark; added cells test depth, branching, list aggregation, relation binding, and state tracking under competence gates. \\
Scale generalization is unresolved & Tables~\ref{tab:llama1b-stress}, \ref{tab:carrier-split}, \ref{tab:field-routing-hardening}, \ref{tab:main-8b-time-prefix}, \ref{tab:8b-no-latent}, \ref{tab:8b-prefix-readout-ablation}, \ref{tab:repair-boundary} & 1B arithmetic-like cells preserve value/KV transfer, but 8B and repaired task cells shift to prompt-prefix/full-cache carriers that appear already before latent rollout and remain sensitive to visible fields at readout. \\
Carrier calls depend on competence gates, thresholds, or tie order & Table~\ref{tab:metric-audit} & The large-$N$ 1B/8B carrier split is unchanged under $.70/.80/.90$ competence gates and stricter sufficiency/competitor thresholds; exact target-score ties are rare in the cache-interchange state-score rows. \\
The late-layer split could be arbitrary & Tables~\ref{tab:gpt2-block-scan}, \ref{tab:scit-bench-v2} & GPT-2 has a stable middle-to-late restricted block, but architecture-relative localization shifts at 1B. \\
Closed candidate sets inflate random baselines & Figure~\ref{fig:candidate-stress}; Tables~\ref{tab:llama1b-stress}, \ref{tab:sampled-freegen} & Enlarged candidate sets and sampled decoding preserve the main exact-context separation; wide-OOD random wins are treated as high-baseline cases and interpreted through continuous margins. \\
The effect may be only hidden-state replacement & Tables~\ref{tab:cache-diagnostic}, \ref{tab:vector-cache}, \ref{tab:simcot-vector-cache}, \ref{tab:component-sufficiency} & Hidden-only and key-only patches are weak; value/cache replacement carries the arithmetic transfer, with checkpoint-dependent readout support. \\
The trajectory could be a last-token, single-head, or position-template trigger & Table~\ref{tab:suffix-window-main}; Appendix Tables~\ref{tab:trajectory-content}, \ref{tab:head-pareto}, \ref{tab:value-geometry}, \ref{tab:noise-prefix}, \ref{tab:value-gradient-readout}, \ref{tab:cross-scale-head-pareto} & Multi-token suffixes are needed, ranked multi-head patches recover transfer much faster than random head sets, cross-template pairs align geometrically, benign prefix noise preserves localization, answer-margin gradients concentrate in the same value region, and held-out cross-scale runs show ranked-head structure in the active carrier. \\
The carrier-shift interpretation may be only verbal & Table~\ref{tab:aux-carrier-diagnostics} & Stable-rank, effective-rank, source-difference norm, attention-entropy, top-head attribution, and time-resolved scans provide auxiliary support for the 8B prompt-prefix carrier; the 1B readout diagnostics are mixed and remain secondary to causal interchange and corruption evidence. \\
The localized region may be globally fragile rather than function-specific & Tables~\ref{tab:necessity}, \ref{tab:cache-ablation-specificity} & Matched corruption is the primary necessity test; source-free zero/mean ablation is a harsher specificity check showing that damaging the same cache spans disrupts recipient readout without constituting counterfactual transfer. \\
Longer horizons or different training paths may change the carrier & Tables~\ref{tab:training-horizon-boundary}, \ref{tab:codi-multiseed-bootstrap}, \ref{tab:fixedrule-support-boundary} & Stronger true-final and range-augmented recipes can recover latent-tail transfer, but adjacent recipes and new CODI-style seeds expose weak or prompt-prefix/full-cache regimes; the carrier is path-, seed-, and support-sensitive. \\
Decoded behavior may not match teacher-forced scores & Tables~\ref{tab:freegen}, \ref{tab:freegen-simcot}, \ref{tab:sampled-freegen}, \ref{tab:freegen-cacheonly} & Greedy and limited sampled decoding preserve the exact-context arithmetic pattern; broader deployment remains outside scope. \\
Prompt-prefix carrier timing is not fully separated & Tables~\ref{tab:8b-no-latent}, \ref{tab:8b-prefix-readout-ablation}, \ref{tab:field-routing-hardening}, \ref{tab:main-8b-time-prefix}, \ref{tab:exprlist-field}, \ref{tab:repair-boundary}, \ref{tab:carrier-split} & No-latent scoring shows the prompt state is already sufficient before rollout; post-prefix ablations show visible fields are still readout-active after latent encoding. This favors visible-field readout but does not rule out early completion. \\
\bottomrule
\end{tabularx}
\caption{Concern-to-evidence map for the appendix. Each row points to compact diagnostic evidence for a likely weakness.}
\label{tab:concern-evidence-map}
\end{table*}

\subsubsection{Carrier Ledger}
Table~\ref{tab:carrier-evidence-ledger}, the compact ledger behind the main carrier-map claims, puts gate, sufficiency, necessity, decoded validation, failed-gate control, and threshold/tie audit in one place. Its purpose is to make the large-model rows answer to the same contract as the GPT-2 arithmetic rows rather than read like an informal stress-test appendix.

\begin{table*}[t]
\centering
\begingroup
\scriptsize
\setlength{\tabcolsep}{2pt}
\renewcommand{\arraystretch}{1.05}
\begin{tabularx}{\textwidth}{@{}L{0.13\textwidth}L{0.13\textwidth}L{0.12\textwidth}L{0.12\textwidth}L{0.13\textwidth}L{0.15\textwidth}L{0.12\textwidth}Y@{}}
\toprule
\textbf{Cell} & \textbf{Breadth} & \textbf{Gate} & \textbf{Active carrier} & \textbf{Oracle suff. TW} & \textbf{Matched corruption} & \textbf{Decoded CF} & \textbf{Call} \\
\midrule
1B arithmetic-like & two-hop, three-hop, conditional & FG 1.00 ($N=192$) & latent-tail KV & 1.00 vs 0.00 ($N=192$) & corrupt 0.00 vs 1.00; dLP 12.79 ($N=384$) & 1.00 vs 0.00 ($N=192$) & latent-tail repl. \\
1B entity lookup & single visible binding & FG pass 1.00 ($N=64$) & prompt-prefix KV & 1.00 vs 0.00 ($N=64$) & corrupt 0.00 vs 1.00; dLP 21.87 ($N=128$) & 0.88 vs 0.00 ($N=64$) & carrier shift \\
1B relation chain & 12-room relation repair & FG 1.00 ($N=48$) & prompt-prefix KV & 1.00 vs 0.00 ($N=48$) & corrupt 0.00 vs 1.00; dLP 20.69 ($N=128$) & 1.00 vs 0.00 ($N=48$) & carrier shift \\
8B arithmetic-like & two-hop, three-hop, conditional & FG 0.99 ($N=120$) & prompt-prefix KV & 1.00 vs 0.00 ($N=120$) & corrupt 0.00 vs 0.99; dLP 13.64 ($N=192$) & 1.00 vs 0.00 ($N=120$) & scale shift \\
8B entity lookup & single visible binding & FG 1.00 ($N=40$) & prompt-prefix KV & 1.00 vs 0.00 ($N=40$) & corrupt 0.00 vs 1.00; dLP 16.25 ($N=64$) & 1.00 vs 0.00 ($N=40$) & carrier shift \\
8B repair/binding & expression, list, relation & FG 1.00 ($N=120$) & prompt-prefix KV & 1.00 vs 0.00 ($N=120$) & corrupt 0.00 vs 1.00; dLP 21.07 ($N=192$) & 1.00 vs 0.00 ($N=120$) & shift \\
Qwen3-4B boundary & same four probes & closed-set 0.25--0.29 & no call & best oracle 0.25--0.33; latent/prefix/full & not interpreted & not run & failed gate \\
Metric audit & 12 large-$N$ cells; 855,200 score rows & fixed rule & threshold/tie check & all 16 threshold settings unchanged & not corruption & exact ties 0.063\%; near 0.072\% & rules audited \\
\bottomrule
\end{tabularx}
\caption{Comprehensive carrier-evidence ledger generated from the retained run196--run198 HPC3 summaries. Gate reports unpatched recipient free-generation accuracy where decoded validation is available. Oracle sufficiency and decoded counterfactual (CF) columns report active versus inactive segment rates. Matched corruption reports recipient target-win after corrupting the active versus inactive segment and the active-segment mean log-probability drop (dLP). The final two rows make failed-gate and reporting-rule audits visible in the same table rather than relegating them to prose.}
\label{tab:carrier-evidence-ledger}
\endgroup
\end{table*}

\subsubsection{8B Field Routing}
Table~\ref{tab:field-routing-hardening} decomposes the 8B prefix-shift row into visible prompt fields rather than leaving it as an opaque full-cache positive. It asks which lookup entries, rule/fact spans, query fields, or support-boundary rows carry the source label. Table~\ref{tab:8b-prefix-readout-ablation} adds the complementary recipient-side question: after prompt encoding or latent rollout, does ablating the same visible-field cache entries disrupt answer readout?

\begin{table*}[t]
\centering
\appendixtablefont
\setlength{\tabcolsep}{3pt}
\begin{tabularx}{\textwidth}{@{}L{0.15\textwidth}L{0.09\textwidth}ccccL{0.22\textwidth}Y@{}}
\toprule
\textbf{8B cell} & \textbf{Split} & \textbf{$N$} & \textbf{Pref.} & \textbf{Lat.} & \textbf{Src. ctrl} & \textbf{Best visible field TW} & \textbf{Call} \\
\midrule
2-hop arithmetic & in-domain & 40 & 1.000 & 0.000 & 1.000 & initial\_clause:1.00; facts\_only:1.00; initial\_red:1.00 & visible source-binding \\
3-hop arithmetic & in-domain & 40 & 1.000 & 0.000 & 1.000 & initial\_clause:1.00; facts\_only:1.00; initial\_red:1.00 & visible source-binding \\
entity lookup & in-domain & 40 & 1.000 & 0.000 & 1.000 & lookup\_table:1.00; matching\_lookup\_entry:1.00; facts\_only:1.00 & visible source-binding \\
relation chain & in-domain & 152 & 1.000 & 0.000 & 1.000 & query:1.00; person\_table:0.00; room\_table:0.00 & visible source-binding \\
expression tree & far-OOD & 81 & 0.988 & 0.000 & 0.988 & facts\_only:0.98; input\_bindings:0.63; left\_inputs:0.59 & visible source-binding \\
expression tree & ultra-OOD & 81 & 0.185 & 0.136 & 0.200 & facts\_only:0.22; input\_bindings:0.20; left\_inputs:0.20 & boundary/no call \\
list filter/sum & far-OOD & 80 & 1.000 & 0.000 & 1.000 & rules\_only:1.00; facts\_only:1.00; threshold\_value:0.44 & visible source-binding \\
list filter/sum & ultra-OOD & 78 & 0.949 & 0.038 & 0.922 & rules\_only:0.93; facts\_only:0.93; threshold\_value:0.26 & visible source-binding \\
\bottomrule
\end{tabularx}
\caption{8B field-routing hardening for prompt-prefix carrier shifts. Prefix (Pref.) and latent (Lat.) columns report oracle cache-only K/V target-win rates; Src. ctrl reports the larger source-label win rate among random and non-oracle prompt-prefix patches. The field column reports the strongest visible prompt spans from the same held-out run. Successful 8B cells are not opaque full-cache wins: lookup entries, facts, rules, or query fields recover transfer while latent-suffix K/V remains inactive. High Src. ctrl is interpreted as visible source-field copying rather than oracle-specific latent reasoning; weak-prefix rows remain no-call boundary cells. Table~\ref{tab:8b-prefix-readout-ablation} gives the inverse readout-sensitivity check on recipient states.}
\label{tab:field-routing-hardening}
\end{table*}

\subsubsection{Metric Audit}
Table~\ref{tab:metric-audit} covers three reporting choices before the mechanism tables: competence gates, carrier thresholds, and tie-breaking. The competence-gate audit uses unpatched recipient rows in the held-out large-$N$ carrier split plus the Qwen3-4B sanity-boundary row. The carrier-threshold sweep uses the same large-$N$ split as Table~\ref{tab:carrier-split}; changing active-carrier floors and competitor ceilings does not change any 1B/8B split call because the relevant target-win rates sit at observed endpoints. The tie audit is computed directly from retained JSONL state-score rows.

\begin{table*}[t]
\centering
\appendixtablefont
\begin{tabularx}{\textwidth}{@{}L{0.25\textwidth}L{0.28\textwidth}Y@{}}
\toprule
\textbf{Audit} & \textbf{Rows / sweep} & \textbf{Outcome} \\
\midrule
Carrier threshold grid & Sufficiency floor $\{.75,.80,.85,.90\}$ crossed with competitor ceiling $\{.10,.20,.25,.30\}$ on $12$ large-$N$ 1B/8B carrier cells & Every grid point gives the same calls: $5$ latent-tail cells and $7$ prompt-prefix cells; no row becomes mixed or inconclusive. \\
Competence gate grid & Competence floor $\{.70,.80,.90\}$ on the same $12$ large-$N$ 1B/8B carrier cells, using unpatched clean closed-set competence & Every grid point keeps all $12$ cells and the same $5$ latent-tail / $7$ prompt-prefix calls; raw competence is $.992$--$1.000$. \\
Low-competence boundary check & Qwen3-4B sanity probes under the same closed-set competence rule & Four sanity probes remain below even a $.70$ gate ($.25$--$.29$), with best oracle target win only $.25$--$.33$; no mechanism call is unlocked by relaxing the default gate. \\
Default carrier threshold & Sufficiency $\geq .80$, competitor $\leq .25$ & The default is a reporting rule tied to a chance-level competitor ceiling; it is not needed to separate the endpoint large-$N$ split rows. \\
Target-win tie audit & $855{,}200$ state-score cache-interchange rows in the retained JSONL outputs & Exact top-score ties occur in $540$ rows ($.063\%$); top-two score gaps within $10^{-3}$ occur in $617$ rows ($.072\%$). \\
\bottomrule
\end{tabularx}
\caption{Metric audit for competence-gate, carrier-threshold, and tie-order reporting risks. The audit is generated from retained summaries and JSONL score rows by \texttt{summarize\_scit\_metric\_audit.py}. It does not replace margins, free generation, or matched corruption for high-baseline cells; it shows that the headline large-$N$ carrier split and cache-interchange target-win rows are not artifacts of the stated competence gate, carrier threshold, or frequent exact ties.}
\label{tab:metric-audit}
\end{table*}

\subsubsection{Method Positioning}
Table~\ref{tab:method-positioning} separates causal interchange evidence from auxiliary probes and decoded validations. It also clarifies how SCIT relates to activation patching, causal tracing, component splits, probes, free-generation validation, and matched corruption.

\begin{table*}[t]
\centering
\appendixtablefont
\begin{tabularx}{\textwidth}{@{}L{0.22\textwidth}L{0.26\textwidth}L{0.24\textwidth}Y@{}}
\toprule
\textbf{Diagnostic} & \textbf{Typical question} & \textbf{SCIT use} & \textbf{Why it matters here} \\
\midrule
Activation patching / causal tracing & Which hidden site changes an output? & Interchange source cache suffixes under exact counterfactual semantics & Keeps the high-level variable fixed by the task rather than inferred after observing a patching effect. \\
Residual hidden-state replacement & Is a single current vector sufficient? & Hidden-only and cache-only baselines & Separates vector-state accounts from accumulated cache-trajectory accounts. \\
Key/value component split & Is transfer routing-side or content-side? & Patch keys, values, K/V together, and matched corruptions & Shows whether the source effect follows attention routing or value content. \\
Linear probes / readout directions & Is a variable readable from a state? & Value probes and probe-direction interventions & Tests content without treating decodability as causal modularity. \\
Free-generation validation & Does a closed-set win survive decoding? & Greedy and sampled patched decoding on headline controls & Guards against teacher-forced candidate-set artifacts. \\
Matched random corruption & Is the localized region needed? & Replace recipient regions with in-distribution random-source regions & Adds a necessity check that is closer to the patch distribution than zero ablation. \\
\bottomrule
\end{tabularx}
\caption{Positioning \method{} relative to standard diagnostics. The contribution is the combined causal contract for latent-CoT rollouts: exact counterfactual source construction, component and segment splits, semantic controls, decoded validation, and matched corruption.}
\label{tab:method-positioning}
\end{table*}

\subsubsection{Evidence Map}
Table~\ref{tab:evidence-map} is a compact reading key for the rest of the appendix: each row names the weakness being addressed and the diagnostic family that bears on it.

\begin{table*}[t]
\centering
\appendixtablefont
\setlength{\tabcolsep}{4pt}
\begin{tabularx}{\textwidth}{@{}L{0.16\textwidth}L{0.33\textwidth}Y@{}}
\toprule
\textbf{Issue} & \textbf{Diagnostic} & \textbf{Takeaway} \\
\midrule
Carrier & Layer-group value-cache interchange on matched counterfactual pairs & In GPT-2-scale checkpoints, value patches in a middle-to-late window transfer the counterfactual answer; nonlate values and keys are weak. \\
Hidden state & Hidden-state-only, cache-only, and key/value separation & The current hidden vector alone is weak; the effect follows the value cache, with checkpoint-dependent readout support from the source hidden state. \\
Semantic slot & Same-answer, partial-variable, and exact-context source controls & Exact counterfactual trajectories dominate semantic controls, arguing against an answer-only or modular intermediate-slot account. \\
Necessity & Matched random corruption of recipient cache entries & Corrupting the same GPT-2 value region selectively damages recipient-answer preservation; the Sim-CoT replication is directional but weaker. \\
Boundary behavior & Greedy/sampled decoding, 1B/8B calibration cells, and targeted task repairs & Arithmetic cells preserve the decoded exact-context pattern, while expression/list, relation-chain, and 8B cells show that the carrier segment can shift or fail OOD. \\
\bottomrule
\end{tabularx}
\caption{Qualitative evidence map. The main text foregrounds the compact numerical evidence in Table~\ref{tab:headline-results}; this table lists the role of the supporting experiments.}
\label{tab:evidence-map}
\end{table*}

\FloatBarrier
\subsection{Core Arithmetic Contract}
\label{app:core-arithmetic}

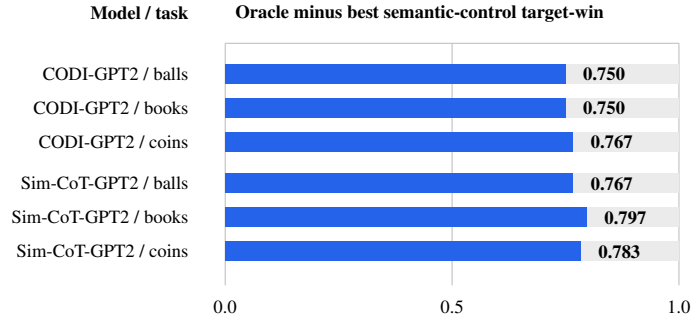
\begin{figure*}[t]
\centering
\begin{tikzpicture}[
barlabel/.style={font=\scriptsize, anchor=east},
barvalue/.style={font=\scriptsize\bfseries, anchor=west},
axislabel/.style={font=\scriptsize, anchor=north}
]
\node[font=\scriptsize\bfseries, anchor=east] at (3.35,0.35) {Model / task};
\node[font=\scriptsize\bfseries, anchor=west] at (3.70,0.35) {Oracle minus best semantic-control target-win};
\draw[black!25] (3.70,-0.05) -- (9.70,-0.05);
\foreach \x/\lab in {3.70/0.0,6.70/0.5,9.70/1.0} {
  \draw[black!25] (\x,-0.05) -- (\x,-3.25);
  \node[axislabel] at (\x,-3.32) {\lab};
}

\node[barlabel] at (3.35,-0.45) {CODI-GPT2 / balls};
\fill[black!8] (3.70,-0.58) rectangle (9.70,-0.32);
\fill[prefixblue] (3.70,-0.58) rectangle (8.20,-0.32);
\node[barvalue] at (8.30,-0.45) {0.750};

\node[barlabel] at (3.35,-0.90) {CODI-GPT2 / books};
\fill[black!8] (3.70,-1.03) rectangle (9.70,-0.77);
\fill[prefixblue] (3.70,-1.03) rectangle (8.20,-0.77);
\node[barvalue] at (8.30,-0.90) {0.750};

\node[barlabel] at (3.35,-1.35) {CODI-GPT2 / coins};
\fill[black!8] (3.70,-1.48) rectangle (9.70,-1.22);
\fill[prefixblue] (3.70,-1.48) rectangle (8.30,-1.22);
\node[barvalue] at (8.40,-1.35) {0.767};

\node[barlabel] at (3.35,-1.90) {Sim-CoT-GPT2 / balls};
\fill[black!8] (3.70,-2.03) rectangle (9.70,-1.77);
\fill[prefixblue] (3.70,-2.03) rectangle (8.30,-1.77);
\node[barvalue] at (8.40,-1.90) {0.767};

\node[barlabel] at (3.35,-2.35) {Sim-CoT-GPT2 / books};
\fill[black!8] (3.70,-2.48) rectangle (9.70,-2.22);
\fill[prefixblue] (3.70,-2.48) rectangle (8.48,-2.22);
\node[barvalue] at (8.58,-2.35) {0.797};

\node[barlabel] at (3.35,-2.80) {Sim-CoT-GPT2 / coins};
\fill[black!8] (3.70,-2.93) rectangle (9.70,-2.67);
\fill[prefixblue] (3.70,-2.93) rectangle (8.40,-2.67);
\node[barvalue] at (8.50,-2.80) {0.783};
\end{tikzpicture}
\caption{Enlarged-candidate semantic-control check. Bars show oracle target-win minus the best non-oracle semantic-control target-win after expanding the scored answer set to about 12--13 candidates. The margins remain large, reducing the concern that semantic-control separation is an artifact of a small four- or five-answer candidate set.}
\label{fig:candidate-stress}
\end{figure*}

Tables~\ref{tab:localization}, \ref{tab:semantic}, and~\ref{tab:necessity} contain the direct numerical support for the local GPT-2 arithmetic claim in the main text: transfer, semantic-source controls, and matched corruption behind Table~\ref{tab:headline-results}. Figure~\ref{fig:candidate-stress} checks that the semantic-control gap is not only a small-candidate-set artifact.

\begin{table*}[t]
\centering
\appendixtablefont
\setlength{\tabcolsep}{3pt}
\begin{tabular}{@{}lccccc@{}}
\toprule
\textbf{Model / Template} & \textbf{All} & \textbf{Late} & \textbf{Late 8--9} & \textbf{Late 10--11} & \textbf{Nonlate} \\
\midrule
CODI-GPT2 / balls & \twostep{1.000}{0.996} & \twostep{1.000}{0.996} & \twostep{0.992}{0.883} & \twostep{0.546}{0.029} & \twostep{0.492}{0.008} \\
CODI-GPT2 / books & \twostep{1.000}{1.000} & \twostep{1.000}{1.000} & \twostep{0.992}{0.875} & \twostep{0.567}{0.017} & \twostep{0.475}{0.008} \\
CODI-GPT2 / coins & \twostep{1.000}{1.000} & \twostep{1.000}{1.000} & \twostep{1.000}{0.908} & \twostep{0.492}{0.017} & \twostep{0.400}{0.000} \\
Sim-CoT-GPT2 / balls & \twostep{0.983}{1.000} & \twostep{0.942}{1.000} & \twostep{0.750}{0.992} & \twostep{0.033}{0.133} & \twostep{0.033}{0.175} \\
Sim-CoT-GPT2 / books & \twostep{0.992}{1.000} & \twostep{0.950}{1.000} & \twostep{0.833}{0.958} & \twostep{0.075}{0.192} & \twostep{0.108}{0.208} \\
Sim-CoT-GPT2 / coins & \twostep{0.975}{1.000} & \twostep{0.942}{1.000} & \twostep{0.792}{0.992} & \twostep{0.017}{0.133} & \twostep{0.067}{0.183} \\
\bottomrule
\end{tabular}
\caption{Exact late-value localization results. Entries are target-win rates at latent steps 5/6 for oracle counterfactual value-suffix patching with the joint cache--state intervention. CODI-GPT2 balls has 240 examples per cell; the other rows have two 60-example seeds. Across rows, step-6 SEM is at most 0.004 for late and 0.037 for nonlate.}
\label{tab:localization}
\end{table*}

The semantic-control block asks whether the same transfer can be explained by sources that match only the final answer, a partial variable, or an intermediate. The key quantity is the oracle-control gap, because closed candidate sets can give nonzero wins to mismatched sources.

\begin{table*}[t]
\centering
\appendixtablefont
\setlength{\tabcolsep}{3pt}
\begin{tabular}{@{}lcccc@{}}
\toprule
\textbf{Model / Template} & \textbf{Oracle} & \textbf{Same CF Ans.} & \textbf{Same Partial} & \textbf{Same Int. / Wrong Var.} \\
\midrule
CODI-GPT2 / balls & \twostep{1.000}{0.988} & \twostep{0.404}{0.400} & \twostep{0.346}{0.325} & \twostep{0.192}{0.175} \\
CODI-GPT2 / books & \twostep{0.992}{0.992} & \twostep{0.437}{0.429} & \twostep{0.367}{0.367} & \twostep{0.117}{0.100} \\
CODI-GPT2 / coins & \twostep{1.000}{1.000} & \twostep{0.450}{0.450} & \twostep{0.308}{0.308} & \twostep{0.125}{0.142} \\
Sim-CoT-GPT2 / balls & \twostep{0.925}{1.000} & \twostep{0.235}{0.353} & \twostep{0.242}{0.375} & \twostep{0.175}{0.225} \\
Sim-CoT-GPT2 / books & \twostep{0.892}{1.000} & \twostep{0.325}{0.483} & \twostep{0.292}{0.433} & \twostep{0.150}{0.217} \\
Sim-CoT-GPT2 / coins & \twostep{0.908}{1.000} & \twostep{0.237}{0.356} & \twostep{0.217}{0.367} & \twostep{0.133}{0.192} \\
\bottomrule
\end{tabular}
\caption{Exact semantic source controls. Entries are target-win rates at latent steps 5/6 for late-value joint cache--state patching. CODI-GPT2 balls aggregates 240 examples per cell; the other rows use two 60-example seeds. Across reported cells, step-6 SEMs are at most 0.046. Because candidate-set scoring can give nonzero wins to random or mismatched sources, the main statistic is the oracle-control margin, not raw distance from nominal chance.}
\label{tab:semantic}
\end{table*}

The enlarged-candidate check repeats the semantic-control separation with a larger answer set, testing whether the oracle-control margin is merely an artifact of four- or five-choice scoring.

The necessity block reverses the intervention direction. Instead of asking whether a source cache is sufficient to transfer the counterfactual answer, it asks whether corrupting the recipient's same cache region selectively damages recipient-answer preservation.

\begin{table}[t]
\centering
\appendixtablefont
\begin{tabular}{@{}llcc@{}}
\toprule
\textbf{Region} & \textbf{Comp.} & \textbf{Target Win} & \textbf{Rec. Drop} \\
\midrule
Late & Value & \twostep{0.767}{0.211} & \twostep{2.431}{13.655} \\
Nonlate & Value & \twostep{1.000}{1.000} & \twostep{0.006}{0.013} \\
Late 8--9 & Value & \twostep{0.906}{0.319} & \twostep{0.874}{9.837} \\
Late 10--11 & Value & \twostep{1.000}{0.986} & \twostep{0.008}{0.186} \\
Late & Key & \twostep{1.000}{1.000} & \twostep{0.002}{0.004} \\
Nonlate & Key & \twostep{1.000}{1.000} & \twostep{0.000}{0.000} \\
\bottomrule
\end{tabular}
\caption{Exact necessity results via matched random suffix corruption on CODI-GPT2, aggregated over balls/books/coins. Entries are target-win rates and recipient-answer log-probability drops at latent steps 5/6; each row aggregates 360 examples.}
\label{tab:necessity}
\end{table}

\subsection{Scope, Generalization, and Repair Boundaries}
\label{app:scope-boundaries}

Tables~\ref{tab:benchmark-scope}--\ref{tab:fixedrule-support-boundary} explain which appendix rows broaden the carrier map and which rows should instead be read as support or competence boundaries. Table~\ref{tab:benchmark-scope} states the intended benchmark scope. The controlled construction is necessary for exact causal interchange, but it deliberately trades task breadth for source-control validity.

\begin{table*}[t]
\centering
\appendixtablefont
\begin{tabularx}{\textwidth}{@{}L{0.18\textwidth}YY@{}}
\toprule
\textbf{Axis} & \textbf{Covered in this diagnostic} & \textbf{Not established} \\
\midrule
Task family & Main diagnostic: one two-hop arithmetic program expressed as balls, books, and coins templates; SCIT-Bench v2 adds clean three-hop arithmetic, conditional arithmetic, entity lookup, graph path, relation chain, finite-state tracking, fixed-rule state updates, expression trees, and list filter/sum as calibration diagnostics; competence-gated repairs expose expression/list, relation-chain, and state-update carrier shifts, plus fixed-rule sequence-holdout failure & Open-domain task families and robust long-horizon state tracking \\
Counterfactual semantics & Exact recipient, oracle, partial-variable, same-answer, and random sources & Naturally occurring benchmark items without constructed source controls \\
Numerical robustness & Wider OOD numeric values with post-transfer and held-fixed counts in 100--500 and transfer counts in 10--99 & Long arithmetic, negatives, carry/borrow-heavy calculations, or symbolic algebra \\
Model scale & Two GPT-2-scale checkpoints plus public/matched LLaMA3-1B calibration diagnostics and matched 1B/8B carrier-split probes, including relation-chain and expression/list repair & A stable latent-tail carrier across large latent-CoT model families \\
Behavioral readout & Teacher-forced target win, greedy and limited temperature-sampling diagnostics, and prompt-only base-model sanity checks & Open-ended downstream deployment behavior or long-horizon reasoning \\
\bottomrule
\end{tabularx}
\caption{Benchmark scope. The SCIT task suite is designed as a controlled causal diagnostic with exact source semantics, not as a general reasoning benchmark.}
\label{tab:benchmark-scope}
\end{table*}

The roadmap makes this scope condition operational by separating the experiments needed for a broad mechanism-generalization claim from those needed for a reliable carrier-regime diagnostic.

\begin{table*}[t]
\centering
\appendixtablefont
\begin{tabularx}{\textwidth}{@{}L{0.18\textwidth}YY@{}}
\toprule
\textbf{Requirement} & \textbf{Needed for a strong generalization claim} & \textbf{Current status in this paper} \\
\midrule
Task structure & A balanced set of exact-counterfactual families: deeper linear arithmetic, branching arithmetic, list aggregation, symbolic binding, path/state tracking, and mixed composition & Positive latent-tail evidence is strongest for arithmetic-like cells; repaired expression/list and relation-chain cells are competent at 8B or 1B/8B but shift to prompt-prefix K/V; minimal and fixed-rule state-update cells also shift to prompt-prefix or remain finite-support bound \\
Model scale & The same diagnostic battery across small, 1B, mid-scale, and 8B-or-larger checkpoints, preferably with at least two architecture families & GPT-2 and 1B arithmetic-like cells support latent-tail value/KV transfer; matched 8B, expression/list repair, and relation-chain probes support full-cache/prompt-prefix transfer but not latent-tail transfer \\
Competence gate & Free-generation and teacher-forced competence before interpreting any causal intervention & Applied throughout the calibration cells; failed or weak cells are treated as competence-limited rather than mechanism failures \\
Carrier contract & For each competent cell, report hidden-only, cache-only, cache-plus-hidden, latent-tail, prompt-prefix, full-cache, key/value/KV, semantic controls, corruption, and free-generation transfer & Implemented for the main arithmetic cells and the current 1B/8B carrier-split probes; repaired expression/list and relation-chain cells have source controls, matched span corruption, prompt-segment or prompt-field split, and patched free generation; minimal and fixed-rule state-update cells have objective, horizon, segment, and sequence-holdout checks \\
Success criterion & Either stable latent-tail value/KV transfer across the grid, or a reproducible map of when the carrier shifts across task and scale & Current evidence supports a limited carrier-regime map, not a universal late-value mechanism \\
\bottomrule
\end{tabularx}
\caption{Generalization roadmap. A full solution to the task/scale objection requires a balanced competence-gated matrix, not additional surface variants of one program.}
\label{tab:generalization-roadmap}
\end{table*}

Table~\ref{tab:relation-surface} tests a narrower question than natural-language generalization: whether the relation-chain carrier call depends on one handcrafted rendering while the executable facts and source--recipient semantics remain fixed.

\begin{table*}[t]
\centering
\appendixtablefont
\setlength{\tabcolsep}{4pt}
\begin{tabularx}{\textwidth}{@{}L{0.23\textwidth}L{0.16\textwidth}L{0.16\textwidth}L{0.20\textwidth}Y@{}}
\toprule
\textbf{Checkpoint / renderings} & \textbf{Recip./oracle FG} & \textbf{Recip./oracle TW} & \textbf{Oracle latent/prompt/full K/V TW} & \textbf{Decision} \\
\midrule
Single-template: original, unseen compact & $48/48,48/48$ & $48/48,48/48$ & $0/1/1$ ($N=48$ each) & prompt-prefix call retained \\
Single-template: unseen brief & $31/48,31/48$ & $37/48,36/48$ & -- & fails competence; scan withheld \\
Multi-template: original, compact, unseen brief & $48/48,48/48$ per form & $48/48,48/48$ per form & $0/1/1$ ($N=48$ each) & prompt-prefix call retained \\
\bottomrule
\end{tabularx}
\caption{Relation-chain surface-form pilot on the same 48 executable worlds under original, compact, and brief renderings. The five competence-passing checkpoint--form cells preserve prompt-prefix rather than latent-suffix transfer, and random prompt-prefix patches select the donor label 48/48 in every passing cell. This is competence-conditional stability under three controlled renderings, not evidence of unrestricted paraphrase, discourse, or strategy robustness.}
\label{tab:relation-surface}
\end{table*}

Table~\ref{tab:scit-bench-v2} bounds the paper's scope. It strengthens three claims: the main arithmetic result survives free decoding, a clean three-hop arithmetic task preserves value-cache transfer, and the public 1B checkpoint supports all-layer value-cache transfer. It also weakens one possible overclaim: restricted layer localization is not stable across architectures, and non-arithmetic cells are not interpretable before the models solve the base task. The ``Best group'' column is the coarse partition used by that calibration run; the controlling evidence for the GPT-2 middle-to-late 8--9 block claim is the contiguous scan in Table~\ref{tab:gpt2-block-scan}.

\begin{table*}[t]
\centering
\begingroup
\setlength{\tabcolsep}{2pt}
\appendixtablefont
\begin{tabular}{@{}llccccc@{}}
\toprule
\textbf{Model} & \textbf{Task cell} & \textbf{Oracle TW} & \textbf{Best group} & \textbf{Margin} & \textbf{Free G/S} & \textbf{Corr. TW} \\
\midrule
CODI-GPT2 & balls/books/coins & 1.000 & q3 09--11 & 0.380--0.463 & 1.000 / 0.978--1.000 & 0.183 \\
Sim-CoT-GPT2 & balls/books/coins & 0.975--1.000 & q3 09--11 & 0.250--0.363 & 1.000 / 1.000 & 0.283--0.367 \\
Sim-CoT-LLaMA3-1B & balls/books/coins & 1.000 & q2 08--11 & 0.412--0.450 & 1.000 / 1.000 & 0.117--0.183 \\
\midrule
CODI-GPT2 & three-hop & 0.863 & q2 06--08 & 0.350 & 0.467 / 0.511 & 0.167 \\
Sim-CoT-GPT2 & three-hop & 0.713 & q2 06--08 & 0.375 & 0.600 / 0.600 & 0.367 \\
Sim-CoT-LLaMA3-1B & three-hop & 1.000 & q3 12--15 & 0.000 & 1.000 / 1.000 & 0.000 \\
Sim-CoT-LLaMA3-1B & conditional & 0.537 & q2 08--11 & 0.262 & 0.033 / 0.056 & 0.383 \\
Current checkpoints & entity/graph & 0.212--0.350 & -- & $\leq 0$ & 0.000 / 0.000 & -- \\
\bottomrule
\end{tabular}
\endgroup
\caption{SCIT-Bench v2 calibration summary at the final latent step. Oracle TW is teacher-forced target win for all-layer value-cache oracle patching with the cache+h intervention. Best group is a coarse calibration-run partition, not the final GPT-2 block-localization result. Margin is oracle minus the best non-oracle semantic source. Free G/S gives greedy/sample parsed-answer success for oracle-patched free generation. Corr. TW is recipient target-win after all-value matched random corruption, where lower values indicate stronger necessity. Core arithmetic rows report ranges over balls, books, and coins. The public 1B three-hop row supports deeper arithmetic transfer but not semantic-source separation, because its semantic controls are not below the oracle. Entity and graph rows for the original current checkpoints fail the competence screen and are not used for mechanism claims.}
\label{tab:scit-bench-v2}
\end{table*}

The augmentation check is competence-gated. The checkpoint is continued on SCIT-Bench v2 training examples, so these cells are not independent evidence of broad generalization; they ask only whether repairing base-task competence is enough to recover the same cache-transfer behavior.

\begin{table*}[t]
\centering
\begingroup
\setlength{\tabcolsep}{3pt}
\appendixtablefont
\begin{tabular}{@{}lccccp{0.33\textwidth}@{}}
\toprule
\textbf{Task cell} & \textbf{Recip. G/S} & \textbf{Oracle G/S} & \textbf{TF Oracle} & \textbf{Margin} & \textbf{Interpretation} \\
\midrule
balls/books/coins & 1.000 / 1.000 & 1.000 / 1.000 & 1.000 & 0.000 & Arithmetic transfer remains saturated, but semantic controls also saturate after augmentation. \\
three-hop & 1.000 / 0.989 & 1.000 / 1.000 & 0.988 & -0.012 & Deeper arithmetic remains transferable; semantic separation is not clean. \\
conditional & 1.000 / 1.000 & 0.333 / 0.333 & 0.863 & -0.025 & Competence is repaired, but free-generation transfer is weak and controls have high baselines. \\
entity lookup & 1.000 / 1.000 & 0.000 / 0.000 & 0.000 & -1.000 & Competence without cache transfer; a clear boundary for the current mechanism claim. \\
graph path & 0.167 / 0.133 & 0.200 / 0.089 & 0.138 & -0.138 & Mostly competence-limited, so not interpretable as a mechanism failure. \\
\bottomrule
\end{tabular}
\endgroup
\caption{Competence-gated SCIT-Bench v2 augmentation on a matched CODI-LLaMA3-1B checkpoint. Recip. G/S is greedy/sample recipient free-generation accuracy. Oracle G/S is greedy/sample success after oracle all-layer value-cache patching with cache+h. TF Oracle is teacher-forced target win for the same oracle patch. Margin is oracle target win minus the best non-oracle semantic source.}
\label{tab:scitaug-boundary}
\end{table*}

The targeted repair runs were added after the broad augmentation check. They do not make the repaired tasks independent benchmark claims, because the checkpoints are continued on the same task families. They answer a narrower concern: when expanded cells are made competent, \method{} still applies and can reveal a different carrier regime.

Expression/list repair gives the strongest in-domain 8B task-breadth extension, and range- and wide-support repairs extend this to larger numeric support. However, the hard-cell replicate shows that the gain is support-bound: far-OOD deployment can be repaired inside the widened support, but ultra-OOD either fails the free-generation gate or remains source-control limited. Because a prompt-only 8B CoT baseline solves the same far-OOD expression-tree split and mostly solves list-filter/sum, this is best read as latent-CODI repair and deployment fragility rather than task infeasibility or an unsuitable base checkpoint. The repaired carrier also shifts to prompt-prefix/full-cache K/V.

Relation-chain supports a prompt-prefix source-answer carrier at 1B/8B scale, while the minimal finite-state and high-update fixed-rule cells localize to visible prompt/source-cache regimes. The large-$N$ relation refresh evaluates both repaired scales with $128$ held-out pairs and the full 12-room candidate vocabulary: oracle prompt-prefix, query-only, and whole-cache K/V patches all reach target win $1.000$, while oracle latent-suffix K/V stays at $0.000$. Random prompt-prefix patches also select the random source label, so this cell is evidence for visible source-field copying, not for a hidden latent-tail reasoning trajectory. The relation-chain field controls decompose this carrier rather than simply repeat the carrier call: query-only patches carry the queried-person binding, while full prefix/cache patches carry the complete source answer under mismatched fields. Hard expression/list variants, OOD relation-chain, graph-path, broader finite-state, and fixed-rule sequence-holdout probes either fail or become source-control limited at the deployment-facing gate, so we treat them as boundary diagnostics rather than broad task-generalization evidence.

\begin{table*}[t]
\centering
\appendixtablefont
\begin{tabularx}{\textwidth}{@{}L{0.22\textwidth}L{0.20\textwidth}L{0.17\textwidth}Y@{}}
\toprule
\textbf{Task} & \textbf{Scale / repair} & \textbf{Gate} & \textbf{Carrier readout} \\
\midrule
Expression/list & 1B targeted repair & Weak in-domain and fails OOD gate & Targeted repair improves teacher-forced scores but does not yield an interpretable mechanism cell. \\
Expression/list & 8B targeted repair & Passes in-domain; OOD fails or remains control-limited & Prompt-prefix/full-cache K/V; latent-tail transfer absent. \\
Relation chain & 1B repaired & Passes held-out free generation; $N=128$ 12-room carrier refresh & Prompt-prefix/query/full-cache K/V; latent-suffix transfer absent. \\
Relation chain & 8B repaired & Passes held-out free generation; $N=128$ 12-room carrier refresh & Same prompt-prefix/query/full-cache field-binding regime at larger scale. \\
Relation chain OOD & 1B repaired & Fails OOD vocab/template gate & Limits repaired-task generalization. \\
Graph path & 1B repaired & Fails short/long hop gates & Path-following competence boundary. \\
Finite-state minimal & 1B tiny objective controls & Passes only one-symbol/one-step cell & Prompt-prefix/full-cache value/KV; segment split points to visible start-state/query. \\
Finite-state tracking & 1B repaired & Fails short/long sequence gates & Broader state-tracking competence boundary. \\
Finite-state tracking & 8B repaired & Fails same/longer sequence gates & Scale alone does not repair the current training recipe. \\
Fixed-rule state & 1B/8B high-update repairs & Fits tiny matched supports only & Prompt-prefix/whole-cache carrier; 8B two-step fails horizon, leave-one, and class-holdout checks. \\
\bottomrule
\end{tabularx}
\caption{Targeted repair boundary tests. Expression/list repair supplies the cleanest in-domain exact-counterfactual task-breadth extension at 8B scale, but it exposes a prompt-prefix/full-cache carrier rather than the arithmetic latent-tail carrier and remains OOD-limited. Repaired relation-chain supplies the main non-arithmetic task-breadth supplement with the same carrier shift; full-vocabulary, sampled-decoding, large-$N$ carrier refreshes, and source-field controls interpret this as visible prompt/source binding. The minimal finite-state and fixed-rule rows are competent only on narrow matched supports and point to prompt binding, not latent state tracking. Rows that fail the free-generation gate or held-out support checks are treated as boundaries rather than broad task-generalization evidence.}
\label{tab:repair-boundary}
\end{table*}

The expression/list block is the most direct response to the task-breadth and scale critique. The important outcome is not that all expanded tasks are solved; the diagnostic distinguishes three regimes: no zero-shot breadth in the existing checkpoints, insufficient 1B targeted repair, and an 8B repair whose causal carrier moves to prompt-prefix/full-cache K/V.

Widening the repair support improves far-OOD deployment, especially for list-filter/sum, but a second hard-cell replicate preserves the same carrier call: latent-suffix patches are ineffective, prompt-prefix/full-cache corruption is damaging, and random-source prompt-prefix patches remain competitive. An enlarged-candidate check with many answer distractors preserves this random-source prompt-prefix effect in the competent hard cells. At ultra-OOD distance, expression-tree becomes a no-call cell and list-filter/sum remains strongly prompt-prefix but source-control limited. The boundary is therefore not attributable to a single teacher-forced artifact, small sample, small candidate set, or obvious path dependence.

\begin{table*}[t]
\centering
\appendixtablefont
\begin{tabularx}{\textwidth}{@{}L{0.18\textwidth}L{0.20\textwidth}L{0.22\textwidth}Y@{}}
\toprule
\textbf{Cell} & \textbf{Gate / stability outcome} & \textbf{Source-control outcome} & \textbf{Interpretation} \\
\midrule
SCIT-Aug 1B/8B baseline & Fails greedy recipient gate & Oracle does not cleanly separate from controls & No zero-shot task-breadth evidence for expression trees or list aggregation. \\
1B targeted repair & Weak in-domain; fails OOD gate & Oracle is weak or tied with controls & Not a positive mechanism cell; competence repair is insufficient at this scale. \\
8B expression-tree repair & In-support and far-OOD sampled gates pass after wide-support repair; ultra-OOD fails & Oracle improves over controls in support, but random prompt-prefix sources remain competitive & Support-distance success with a prompt-prefix carrier, not robust OOD generalization. \\
8B list-filter/sum repair & In-support and far-OOD gates pass; ultra-OOD is partial & Oracle separates in support, while far/ultra-OOD are limited by strong random-source prompt-prefix patches & Same boundary: repair gives wider support competence but not a broad task-generalization claim. \\
\bottomrule
\end{tabularx}
\caption{Expression-tree and list-filter/sum repair summary. The two tasks have exact counterfactual semantics but differ from the original two-hop arithmetic templates: expression-tree uses two intermediate branches before a composed final answer, and list-filter/sum uses threshold selection plus aggregation. The 8B in-support and far-OOD cells pass recipient and oracle-patched free generation after wide-support repair, but segment tests and matched corruption localize the effective carrier to prompt-prefix/full-cache K/V rather than latent-tail K/V. A second hard-cell replicate and enlarged-candidate check preserve this prompt-prefix/source-answer pattern while showing that ultra-OOD remains a competence or source-control boundary.}
\label{tab:exprlist-repair}
\end{table*}

The prompt-only sanity check for expression/list does not use CODI repair or cache patching, so it is not mechanism evidence. It calibrates whether the repaired tasks should be described as intrinsically difficult. The answer is mixed: answer-only prompting is weak, longer scratchpad prompting makes expression-tree robust at 1B and 8B, and list-filter/sum remains weak at 1B and partial-to-strong at 8B. This is why we use the repaired expression/list cells as controlled carrier-regime diagnostics rather than as benchmark improvements.

\begin{table*}[t]
\centering
\appendixtablefont
\begin{tabularx}{\textwidth}{@{}L{0.19\textwidth}L{0.24\textwidth}L{0.24\textwidth}Y@{}}
\toprule
\textbf{Prompt-only baseline} & \textbf{Expression tree} & \textbf{List filter/sum} & \textbf{Paper use} \\
\midrule
1B base, answer-only & Fails in-domain and OOD & Fails in-domain and OOD & Numeric answer-only readout is a strict deployment setting, not a strong natural baseline. \\
8B base, answer-only & Fails in-domain and OOD & Partial but not clean & Same readout caveat; this does not by itself prove the task family is difficult. \\
1B base, CoT prompt & Strong in-domain and OOD with a longer output cap & Weak to partial & Expression-tree is feasible with a scratchpad, so the repair result should not be framed as natural task difficulty. \\
8B base, CoT prompt & Strong in-domain, OOD, and far-OOD with a longer output cap & Partial to strong, including far-OOD & Ordinary scratchpad readout can solve these task families; expression/list is best used to test carrier shifts under latent-CODI repair. \\
\bottomrule
\end{tabularx}
\caption{Prompt-only base-model sanity check for expression-tree and list-filter/sum. The table summarizes held-out base LLaMA 1B/8B runs with an answer-only numeric prompt and a scratchpad CoT prompt; the CoT rows include longer-output sanity checks, an additional seed for the 8B cells, and a farther 8B numeric-range calibration. It is a task-feasibility calibration, not a cache-intervention result.}
\label{tab:exprlist-base}
\end{table*}

The field split asks what the prompt-prefix carrier contains in the repaired 8B expression/list cells. This is not a new positive mechanism claim; it diagnoses the carrier shift by showing that successful prompt-prefix patches move source information through changed problem fields, whereas latent-suffix and answer-prefix spans remain ineffective.

\begin{table*}[t]
\centering
\appendixtablefont
\begin{tabularx}{\textwidth}{@{}L{0.20\textwidth}L{0.27\textwidth}L{0.24\textwidth}Y@{}}
\toprule
\textbf{Cell} & \textbf{Active prompt fields} & \textbf{Inactive fields} & \textbf{Interpretation} \\
\midrule
Expression tree, train/far support & Fact spans and changed input bindings, especially the left-branch inputs & Rule-only, query, answer-prefix, and latent-suffix spans & Transfer is tied to visible source operands rather than to a latent-tail trajectory. \\
Expression tree, ultra-OOD & The same fact/input ordering weakens with the failed deployment gate & Latent suffix remains ineffective & The field split mirrors the ultra-OOD competence boundary rather than rescuing it. \\
List filter/sum, train/far/ultra & Fact/rule spans, with threshold/filter fields as partial carriers & Raw value bindings, answer-prefix, query, and latent-suffix spans & Transfer follows the visible filtering rule that determines which items enter the aggregation. \\
\bottomrule
\end{tabularx}
\caption{Prompt-field ablation for the repaired 8B expression/list cells. The table reports gate-level qualitative outcomes to avoid another dense numeric grid. The active 8B carrier decomposes into visible prompt fields: expression-tree transfer follows source operand/fact bindings inside support, while list-filter/sum follows rule/fact and threshold/filter fields. Latent-suffix and answer-prefix spans do not support transfer.}
\label{tab:exprlist-field}
\end{table*}

The fixed-rule block addresses a distinct scale-and-task generalizability concern. The important comparison is not the matched 8B two-step score by itself, but whether the same repair executes the rule when a small part of the same-horizon support is removed. It does not: seen supports remain competent, while held-out supports fail the free-generation gate. This turns the fixed-rule experiment into a support-generalization boundary rather than an additional positive mechanism claim.

\begin{table*}[t]
\centering
\appendixtablefont
\begin{tabularx}{\textwidth}{@{}L{0.22\textwidth}L{0.25\textwidth}L{0.22\textwidth}Y@{}}
\toprule
\textbf{Probe} & \textbf{Held-out structure} & \textbf{Deployment result} & \textbf{Interpretation} \\
\midrule
Matched 8B two-step repair & No same-horizon sequence holdout & Passes the matched two-step gate; fails shorter/longer horizon checks & A useful competent calibration cell, but its carrier is visible start/query and prompt-prefix cache rather than latent-tail rule execution. \\
Replicated sequence holdout & Three separate splits, each removing multiple bigrams & Seen supports pass; held-out supports fail & The matched success is not stable once small same-horizon regions are removed. \\
Leave-one bigram grid & Train on eight of nine two-symbol sequences & All seen splits pass; every held-out split fails the free-generation gate & Rules out the objection that the earlier held-out splits removed too much support. Partial cells remain below the deployment gate. \\
Net-displacement class holdout & Remove same-net or reverse bigram classes together & Seen supports pass; held-out classes collapse & The partial leave-one signals are best explained by local interpolation from nearby seen bigrams, not by algorithmic state updating. \\
\bottomrule
\end{tabularx}
\caption{Fixed-rule support-generalization boundary for the high-update 8B two-step repair. The table intentionally reports gate-level outcomes rather than another dense numeric grid: the conclusion is qualitative and stable across the leave-one and class-holdout variants.}
\label{tab:fixedrule-support-boundary}
\end{table*}

\subsection{Trajectory Content and Auxiliary Carrier Checks}
\label{app:trajectory-aux}

Tables~\ref{tab:trajectory-content}--\ref{tab:codi-multiseed-bootstrap} ask what the arithmetic trajectory looks like after the causal carrier has already been identified. These diagnostics are secondary: they test content, trajectory length, head concentration, representation alignment, readout sensitivity, and training-path sensitivity, but they do not replace interchange, semantic controls, matched corruption, or free generation as primary evidence. Table~\ref{tab:trajectory-content} summarizes this distinction.

\begin{table*}[t]
\centering
\appendixtablefont
\begin{tabularx}{\textwidth}{@{}L{0.24\textwidth}L{0.25\textwidth}L{0.22\textwidth}Y@{}}
\toprule
\textbf{Probe} & \textbf{What it tests} & \textbf{Observed pattern} & \textbf{Interpretation} \\
\midrule
Block-wise value probes & Whether value blocks contain readable arithmetic variables & Input/context variables are strongly decodable; intermediate and outcome variables are only partial & The trajectory contains task information, but not a modular symbolic trace. \\
Probe-direction intervention & Whether a decoded variable direction is a causal control direction & Probe margins can be moved strongly, but the intended counterfactual answer does not overtake the recipient answer & Decodability is not enough for causal modularity. \\
Semantic source controls & Whether matching an answer or partial variable is sufficient for transfer & Exact counterfactual context dominates same-answer and partial-variable controls & The causal carrier is context-bound rather than a reusable answer or intermediate slot. \\
Cross-template sources & Whether exact context transfer is tied to surface wording & Exact arithmetic contexts transfer across balls/books/coins wording & The trajectory is not merely lexical-template binding. \\
Suffix-window, head, and geometry scans & Whether transfer is a last-token, single-head, or purely lexical trigger & Multi-token suffixes are needed; ranked multi-head patches beat random; exact cross-template pairs align more than off-diagonal pairs & The causal effect is accumulated, sparse-distributed, and partly representation-aligned over the value trajectory. \\
\bottomrule
\end{tabularx}
\caption{Qualitative trajectory-content checks. The table summarizes probe and source-control evidence without presenting the appendix as a separate probing benchmark. The main conclusion is negative but useful: the value-cache trajectory carries readable arithmetic information, yet the readable axes do not form independently transplantable symbolic variables, and the transfer is not explained by a single final token or a single value head.}
\label{tab:trajectory-content}
\end{table*}

Table~\ref{tab:suffix-window-main} asks whether the effect is only a final-token trigger. It is not: CODI needs a short cache window, and Sim-CoT still benefits from more than one latent position, so the operative object is a suffix trajectory rather than a single cache entry.

\begin{table}[t]
\centering
\appendixtablefont
\setlength{\tabcolsep}{3pt}
\begin{tabular}{@{}lcccc@{}}
\toprule
\textbf{Model} & \textbf{$k=0$} & \textbf{$k=1$} & \textbf{$k=2$} & \textbf{$k\geq3$} \\
\midrule
CODI & 0.000 & 0.203 & 0.203 & 0.984--1.000 \\
Sim-CoT & 0.026 & 0.516 & 0.995 & 1.000 \\
\bottomrule
\end{tabular}
\caption{Final-step suffix-window target-win rates for oracle cache+h patches, aggregated over balls/books/coins with 192 examples per row. $k$ is the number of latent cache positions replaced; $k=0$ applies no cache-tail replacement and therefore isolates current-hidden replacement under the cache+h variant. CODI denotes CODI-GPT2; Sim-CoT denotes Sim-CoT-GPT2.}
\label{tab:suffix-window-main}
\end{table}

Table~\ref{tab:adaptive-stop} applies reverse-aligned interchange after independently realized source and recipient stops. The stopping rule was frozen before the held-out intervention slice and did not consult gold labels or correctness.

\begin{table}[t]
\centering
\appendixtablefont
\setlength{\tabcolsep}{4pt}
\begin{tabular}{@{}lccc@{}}
\toprule
\textbf{Patch} & \textbf{All /48} & \textbf{Equal /21} & \textbf{Unequal /27} \\
\midrule
Full latent K/V+$h$ & 47 & 21 & 26 \\
$k=2$ latent K/V+$h$ & 47 & 21 & 26 \\
$k=1$ latent K/V+$h$ & 44 & 20 & 24 \\
Full latent V+$h$ & 47 & 21 & 26 \\
Full latent K+$h$ & 18 & 8 & 10 \\
Full latent K/V only & 27 & 10 & 17 \\
$h$ only & 14 & 5 & 9 \\
Prompt K/V only & 0 & 0 & 0 \\
\bottomrule
\end{tabular}
\caption{Adaptive-stop exact-oracle target-win successes at the frozen clean recipient stop. Among 96 trajectories, realized stops are $T=3/4/5$ for $35/27/34$ trajectories; 27/48 source--recipient pairs have unequal stops. Reverse-aligned full and $k=2$ latent K/V+$h$ each succeed on 47/48 pairs, but K/V-only and $h$-only fail the sufficiency gate. Value+$h$ success does not establish a value-specific carrier because key+$h$ exceeds the inactivity ceiling. The pilot supports joint-object sufficiency, not constituent necessity or intervention-induced re-halting.}
\label{tab:adaptive-stop}
\end{table}

Table~\ref{tab:head-pareto} asks whether the value carrier is a single-head effect or a diffuse all-head effect. Ranked heads recover transfer far faster than random same-size head sets, supporting a sparse-distributed carrier without claiming a named single-head circuit.

\begin{table}[t]
\centering
\appendixtablefont
\setlength{\tabcolsep}{3pt}
\begin{tabular}{@{}lcccc@{}}
\toprule
\textbf{Patch} & \textbf{1h} & \textbf{3h} & \textbf{5h} & \textbf{8h} \\
\midrule
Ranked, cache+h & 0.009 & 0.255 & 0.833 & 0.982 \\
Random, cache+h & 0.000 & 0.003 & 0.005 & 0.023 \\
Ranked, cache-only & 0.023 & 0.287 & 0.824 & 0.968 \\
Random, cache-only & 0.002 & 0.005 & 0.011 & 0.020 \\
\bottomrule
\end{tabular}
\caption{Value-head Pareto diagnostic for CODI-GPT2 arithmetic at the final latent step. Entries are target-win rates after patching the top-ranked or random value heads in the late value window, aggregated over balls/books/coins. Single heads are not sufficient, but a small ranked set recovers the transfer far faster than random same-size sets. The effect is therefore sparse-distributed across value heads rather than a named single-head circuit or a uniform all-head effect.}
\label{tab:head-pareto}
\end{table}

Tables~\ref{tab:cross-scale-head-pareto} and~\ref{tab:aux-carrier-diagnostics} repeat the carrier-shape question across the scale and task cells used in the carrier-regime map. They are secondary support for the segment calls; the causal evidence still comes from interchange, corruption, and decoded behavior.

\begin{table*}[t]
\centering
\appendixtablefont
\setlength{\tabcolsep}{4pt}
\begin{tabular}{@{}llccccc@{}}
\toprule
\textbf{Cell} & \textbf{Active carrier} & \textbf{Top-1} & \textbf{Top-4} & \textbf{Top-8} & \textbf{Top-16} & \textbf{Random-64} \\
\midrule
1B arithmetic & Latent-tail value & 0.000 & 0.727 & 1.000 & 1.000 & 0.445 \\
1B entity & Prompt-prefix K/V & 0.023 & 0.781 & 1.000 & 1.000 & 0.531 \\
8B arithmetic & Prompt-prefix K/V & 0.996 & 1.000 & 1.000 & 1.000 & 0.254 \\
8B entity & Prompt-prefix K/V & 0.000 & 0.504 & 0.996 & 1.000 & 0.117 \\
8B expression & Prompt-prefix K/V & 0.453 & 0.793 & 0.883 & 0.965 & 0.160 \\
8B list & Prompt-prefix K/V & 0.027 & 0.527 & 0.734 & 0.996 & 0.141 \\
\bottomrule
\end{tabular}
\caption{Cross-scale ranked-head Pareto diagnostic on the active carrier for each cell. Entries are final-step target-win rates under cache-only patches on two held-out evaluation replicates ($N=256$ pooled per row), using head rankings selected on a separate small scan set. The active carrier is not uniformly spread across all heads: ranked heads recover transfer much faster than random head sets. Some 8B prompt-prefix cells are highly concentrated, while list-filter/sum remains more distributed and reaches full recovery only at larger ranked-head budgets.}
\label{tab:cross-scale-head-pareto}
\end{table*}

\begin{table*}[t]
\centering
\appendixtablefont
\setlength{\tabcolsep}{3pt}
\begin{tabularx}{\textwidth}{@{}L{0.16\textwidth}L{0.18\textwidth}L{0.18\textwidth}L{0.16\textwidth}L{0.16\textwidth}Y@{}}
\toprule
\textbf{Cell} & \textbf{SCIT carrier} & \textbf{$\Delta$KV stable rank, Lat./Pref.} & \textbf{$\Delta$KV norm, Lat./Pref.} & \textbf{Top-8 attn., Lat./Pref.} & \textbf{Time scan} \\
\midrule
1B arithmetic & Latent-tail value/KV & 12.1 / 11.8 & 128 / 131 & .111 / .781 & Latent c-only 4/6; latent +h 6/6. Prefix c-only 0/6. \\
8B arithmetic & Prompt-prefix K/V & 36.9 / 12.4 & 5.9 / 320 & .011 / .973 & Prefix c-only/+h 6/6; latent 0/6. \\
8B expression & Prompt-prefix K/V & 33.9 / 15.3 & 12.9 / 418 & .006 / .978 & Prefix c-only/+h 6/6; latent 0/6. \\
8B list & Prompt-prefix K/V & 40.7 / 30.7 & 39.2 / 374 & .013 / .949 & Prefix c-only/+h 6/6; latent 0/6. \\
8B relation & Prompt-prefix K/V & 20.9 / 35.3 & 141 / 304 & .033 / .959 & Prefix c-only/+h 6/6; latent 0/6. \\
\bottomrule
\end{tabularx}
\caption{Auxiliary carrier-shift diagnostics. $\Delta$KV is the source-minus-recipient K/V matrix at the final latent step; stable rank is $\|M\|_F^2/\|M\|_2^2$, and norm is Frobenius norm. The full diagnostic file also reports centered effective rank and attention-entropy bootstrap intervals; these are summarized here only through the compact stable-rank, norm, and readout-mass columns. Top-8 attention reports answer-readout mass to the latent suffix and prompt prefix for heads ranked by an independent carrier-head scan. The time scan counts the number of latent steps, out of six, where oracle K/V patching reaches saturated target-win in a required-coverage run that compares cache-only and cache+h patches; Table~\ref{tab:main-8b-time-prefix} gives the later $N=48$ four-cell follow-up. These diagnostics support the prompt-prefix carrier in the 8B cells, but they are not causal proof by themselves. The 1B row is intentionally mixed: SCIT patches identify a latent-tail carrier, while answer-readout attention remains prompt-heavy and prompt-prefix cache+h has isolated hidden-state-assisted positives.}
\label{tab:aux-carrier-diagnostics}
\end{table*}

Tables~\ref{tab:value-geometry}--\ref{tab:cache-ablation-specificity} test artifact explanations that could otherwise make the arithmetic localization look stronger than it is: lexical alignment, absolute position or template dependence, readout-sensitivity mismatch, and generic cache fragility. These are robustness and specificity checks, not standalone causal proofs.

\begin{table}[t]
\centering
\appendixtablefont
\setlength{\tabcolsep}{3pt}
\begin{tabular}{@{}lcccc@{}}
\toprule
\textbf{Block} & \textbf{CKA} & \textbf{Pair} & \textbf{Offdiag} & \textbf{Gap} \\
\midrule
Nonlate & .999 & .996 & .720 & .277 \\
Late 8--9 & .999 & .995 & .563 & .432 \\
Late 10--11 & .999 & .996 & .432 & .564 \\
\bottomrule
\end{tabular}
\caption{Cross-template value-geometry check at the final latent step. Rows average balls/books/coins pairs matched by the same arithmetic state. CKA is saturated and should not be over-interpreted; the useful signal is the paired-minus-offdiagonal cosine gap, which is larger in the value blocks implicated by the SCIT transfer than in the nonlate block. This is representation corroboration, not causal proof.}
\label{tab:value-geometry}
\end{table}

\begin{table}[t]
\centering
\appendixtablefont
\setlength{\tabcolsep}{3pt}
\begin{tabular}{@{}lccccc@{}}
\toprule
\textbf{Model} & \textbf{All} & \textbf{Late} & \textbf{8--9} & \textbf{10--11} & \textbf{Nonlate} \\
\midrule
CODI & 1.000 & 0.996 & 0.917 & 0.017 & 0.000 \\
Sim-CoT & 1.000 & 1.000 & 0.984 & 0.142 & 0.134 \\
\bottomrule
\end{tabular}
\caption{Benign prefix-noise robustness. Entries are final-step target-win rates averaged over balls/books/coins after prepending a neutral natural-language sentence to the prompt. The same late and middle-to-late value-cache localization remains active, while final-only and nonlate patches stay weak. This argues against an absolute-position or rigid-template explanation of the arithmetic localization.}
\label{tab:noise-prefix}
\end{table}

\begin{table}[t]
\centering
\appendixtablefont
\setlength{\tabcolsep}{3pt}
\begin{tabular}{@{}lccc@{}}
\toprule
\textbf{Task} & \textbf{L9H1} & \textbf{L8H1} & \textbf{L8H8} \\
\midrule
Balls & 1.225 & 1.060 & 0.865 \\
Books & 1.428 & 1.167 & 0.914 \\
Coins & 1.305 & 1.102 & 0.915 \\
\bottomrule
\end{tabular}
\caption{Value-gradient readout diagnostic for CODI-GPT2 under late-value patching. Entries are average gradient-times-value magnitudes for the answer-margin objective at recurring high-sensitivity heads. Sensitivity concentrates in the same layers 8--9 value region implicated by suffix interchange. This is a readout-sensitivity check, not proof of a single-head circuit.}
\label{tab:value-gradient-readout}
\end{table}

\begin{table}[t]
\centering
\appendixtablefont
\setlength{\tabcolsep}{3pt}
\begin{tabular}{@{}llcc@{}}
\toprule
\textbf{Span} & \textbf{Ablation} & \textbf{Rec. win} & \textbf{Rec. drop} \\
\midrule
Latent value & zero & 0.425 & 11.43 \\
Latent K/V & zero & 0.388 & 12.49 \\
Latent key & zero & 0.900 & 1.02 \\
Prompt K/V & zero & 0.254 & 13.81 \\
Whole K/V & mean & 0.279 & 30.17 \\
\bottomrule
\end{tabular}
\caption{Source-free cache-ablation specificity check on CODI-GPT2 arithmetic. Entries average balls/books/coins at the final latent step. Direct zero or mean replacement is harsher and more out-of-distribution than matched random corruption, so it is not the primary necessity evidence. It nevertheless shows that damaging value/K/V cache spans disrupts recipient-answer readout much more than latent-key-only zeroing, while broad prompt or whole-cache ablation unsurprisingly causes larger global damage.}
\label{tab:cache-ablation-specificity}
\end{table}

Table~\ref{tab:training-horizon-boundary} changes the training path, latent horizon, and numeric support to test whether latent-tail transfer appears automatically with more training or longer rollouts. It does not: stronger and range-augmented recipes can recover the arithmetic carrier, but adjacent recipes, competent seed replicates, and farther-OOD cells expose carrier-basin and deployment boundaries.

\begin{table*}[t]
\centering
\appendixtablefont
\setlength{\tabcolsep}{4pt}
\begin{tabularx}{\textwidth}{@{}L{0.23\textwidth}L{0.23\textwidth}L{0.22\textwidth}Y@{}}
\toprule
\textbf{Probe} & \textbf{Positive signal} & \textbf{Latent-tail result} & \textbf{Interpretation} \\
\midrule
Quick GPT-2 re-training dynamics & Oracle/full-cache target win rises from 0.58 at checkpoint 100 to 0.81 at checkpoint 440; prompt-prefix tracks the same curve & Latent-tail-only stays weak, roughly 0.14--0.23 target win with near-zero margin & This quick path does not show a clean ontogeny in which prompt binding compresses into a latent-tail carrier. \\
Stronger true-final $T=6$ recipe & Whole-cache and oracle transfer are high; a new-seed replicate preserves the effect & Latent-tail-plus-hidden recovers most oracle transfer, and suffix windows rise smoothly from near zero to high transfer & The arithmetic latent-tail regime is recoverable under stronger training, but it is not automatic. \\
Adjacent $T=6$ recipes & Higher learning rate and wider numeric support keep whole-cache transfer high & Latent-tail suffix patches remain weak while prompt-prefix/full-cache transfer dominates & Carrier choice is training-recipe sensitive even within GPT-2 arithmetic. \\
Stronger true-final $T=12$ & Whole-cache replacement remains strong & Sliding suffix patches from $k=0$ to $k=11$ stay near chance, while prompt-prefix transfer is high & Longer latent rollout alone does not imply latent-tail accumulation. \\
Stronger true-final $T=16$ & Oracle and whole-cache transfer are near saturated; a larger replicate preserves the effect & Latent-tail-plus-hidden is near saturated, and suffix windows rise gradually from one-token failure to full-suffix recovery & Longer horizons can recover a trajectory regime, but only for some recipes; horizon scaling is not monotonic. \\
Range-augmented $T=6$ support-wide & Recipient and oracle free-generation pass on the expanded numeric support across a larger evaluation replicate & Latent-tail-plus-hidden and full-cache transfer are saturated, while prompt-prefix remains weak; prompt-field splits localize the transfer to latent-suffix value/KV, not visible facts & Numeric support expansion can repair the old wide-range boundary inside support and restore the arithmetic latent-tail carrier. \\
Range-augmented $T=16$ support-wide & Recipient/oracle free-generation and closed-set target-win pass on balls/books/coins across additional training seeds and intermediate checkpoints & One seed reproduces latent-tail/full-cache transfer from the earliest scanned checkpoint, while another competent seed is already prompt-prefix/facts/full-cache dominated and leaves the latent suffix ineffective & Long-horizon range augmentation repairs support competence, but the trained-checkpoint carrier basin is seed-sensitive and appears early; early and final prompt-field attribution explain the shifted seed as visible-facts binding. \\
Range-augmented farther OOD & $T=6$ free generation fails and closed-set transfer is near the high baseline; the original $T=16$ run emits off-candidate numeric answers although closed-set target-win remains high & $T=6$ does not separate cleanly; the original $T=16$ run keeps latent-tail/full-cache teacher-forced transfer, but new $T=16$ seeds do not replicate robust far-OOD transfer & Beyond support, internal closed-set energy and generated answers can diverge; this is not robust arithmetic extrapolation. \\
\bottomrule
\end{tabularx}
\caption{Exploratory training-path, latent-horizon, and numeric-support probes. These newly trained GPT-2 CODI checkpoints are not used to replace the original CODI-GPT2 result. They calibrate its scope: stronger training and range augmentation can recover a latent-tail trajectory in selected arithmetic recipes, but adjacent recipes, one competent $T=16$ seed replicate, and farther OOD cells expose prompt-prefix/facts, weak-transfer, or deployment-readout boundary regimes. The T16 seed-replicate dynamics favor early carrier-basin bifurcation over either method instability or a universal late-training compression into latent-tail values.}
\label{tab:training-horizon-boundary}
\end{table*}

\begin{table*}[t]
\centering
\appendixtablefont
\setlength{\tabcolsep}{4pt}
\begin{tabular}{@{}lcccccc@{}}
\toprule
\textbf{Seed} & \textbf{Clean recipient TW} & \textbf{Late 8--9 suff.} & \textbf{Late 10--11 suff.} & \textbf{All-value suff.} & \textbf{Late 8--9 corrupt TW} & \textbf{Drop} \\
\midrule
s204a & .951 [.927,.971] & .138 [.104,.172] & .328 [.281,.375] & .464 [.414,.510] & .904 [.875,.932] & .047 \\
s204b & .938 [.911,.961] & .049 [.029,.070] & .143 [.109,.177] & .214 [.172,.253] & .849 [.812,.883] & .089 \\
s204c & .987 [.974,.997] & .135 [.102,.169] & .516 [.464,.562] & .479 [.427,.526] & .961 [.940,.979] & .026 \\
\bottomrule
\end{tabular}
\caption{CODI-GPT2 multi-seed replication audit. Rows pool balls/books/coins with $N=384$ examples per seed and report bootstrap 95\% intervals in brackets. Sufficiency columns are oracle value-cache target-win rates under the layer-suffix intervention; corruption target win is recipient-answer preservation after matched random late-8--9 value corruption. A follow-up latent/prompt/whole-cache K/V scan gives $.370/.203/.870$ (s204a), $.208/.427/.865$ (s204b), and $.427/.141/.906$ (s204c). Whole-cache transfer is high, but neither disjoint segment reaches the $.80$ local-carrier gate, so the protocol returns no localized carrier and does not run finalist semantic or decoded tests. These competent seeds are therefore seed-sensitivity evidence, not positive replications of the headline CODI-GPT2 mechanism.}
\label{tab:codi-multiseed-bootstrap}
\end{table*}

\subsection{Carrier-Regime and Scale Details}
\label{app:carrier-scale}

Tables~\ref{tab:carrier-split}--\ref{tab:llama1b-stress} give the segment-level evidence behind the carrier-regime map in the main text. They separate latent-tail, prompt-prefix, and whole-cache replacement, then add block scans and OOD/1B checks needed to interpret scale and numeric-support claims. Table~\ref{tab:carrier-split} asks whether the transferable source information is carried by the latent-tail cache or by the prompt-prefix cache. The augmented 1B/8B arithmetic, conditional, and entity rows are refreshed with $N=128$ held-out pairs per task; the repaired relation-chain rows are refreshed with $N=128$ per scale and the full 12-room candidate vocabulary. The component split shows the augmented arithmetic/entity/relation pattern: the matched 1B checkpoint keeps the arithmetic-like signal in the latent tail, while entity lookup and repaired relation-chain move with the prompt prefix.

Table~\ref{tab:carrier-split} adds expression/list repair and the minimal finite-state control, which follow the same prompt-prefix regime when they pass the gate; Table~\ref{tab:exprlist-field} further decomposes the expression/list prompt-prefix carrier into the source fields that changed. In the relation-chain refresh, prompt-prefix, query-only, and whole-cache K/V patches reach $1.000$ target win at both 1B and 8B scale, while latent-suffix K/V remains at $0.000$. The matched 8B checkpoint gives the scale boundary: full-cache transfer is present on every competent cell, but the source signal is already available in the prompt prefix and is not recovered from the latent-tail K/V cache alone.

A matched random segment-corruption follow-up gives the inverse necessity pattern for the same competent cells: 1B arithmetic-like and conditional prompts are damaged by latent-tail value/K/V corruption but not by prompt-prefix corruption, whereas 1B entity lookup, 8B expression/list repair, and the matched 8B cells are damaged by prompt-prefix value/K/V corruption but not by latent-tail corruption. We keep graph path out of the carrier call because its decoded competence remains weak and the corruption readout is correspondingly unstable. The component split therefore reflects a carrier-segment shift rather than only a sufficiency artifact.

\begin{table*}[t]
\centering
\begingroup
\setlength{\tabcolsep}{3pt}
\appendixtablefont
\begin{tabular}{@{}llcccc@{}}
\toprule
\textbf{Checkpoint} & \textbf{Task cell} & \textbf{Latent-tail KV} & \textbf{Prompt-prefix KV} & \textbf{Full KV} & \textbf{Call} \\
\midrule
1B + SCIT aug. & balls/books/coins & 1.000 & 0.000 & 1.000 & latent tail \\
1B + SCIT aug. & three-hop & 1.000 & 0.000 & 1.000 & latent tail \\
1B + SCIT aug. & conditional & 1.000 & 0.000 & 1.000 & latent tail \\
1B + SCIT aug. & entity lookup & 0.000 & 1.000 & 1.000 & prompt prefix \\
1B relation repair & relation chain & 0.000 & 1.000 & 1.000 & prompt prefix \\
1B finite-state tiny & one-symbol/one-step & 0.000 & 1.000 & 1.000 & prompt prefix \\
\midrule
8B + SCIT aug. & balls/books/coins & 0.000 & 1.000 & 1.000 & prompt prefix \\
8B + SCIT aug. & three-hop & 0.000 & 1.000 & 1.000 & prompt prefix \\
8B + SCIT aug. & conditional & 0.000 & 1.000 & 1.000 & prompt prefix \\
8B + SCIT aug. & entity lookup & 0.000 & 1.000 & 1.000 & prompt prefix \\
8B expression/list repair & expression tree & 0.000 & 1.000 & 1.000 & prompt prefix \\
8B expression/list repair & list filter/sum & 0.000 & 1.000 & 1.000 & prompt prefix \\
8B relation repair & relation chain & 0.000 & 1.000 & 1.000 & prompt prefix \\
\bottomrule
\end{tabular}
\endgroup
\caption{Carrier-split diagnostic at latent step 6. Entries are teacher-forced target-win rates for K/V cache replacement by segment. The augmented 1B/8B arithmetic, conditional, and entity rows use large-$N$ held-out refreshes ($N=128$ per task). The repaired relation-chain rows use $N=128$ per scale with the full 12-room candidate vocabulary; query-only K/V patches also reach $1.000$ target win, confirming a visible query/source-field carrier. For the 1B augmented checkpoint, value-only latent-tail patches are also sufficient for balls/books/coins/three-hop and near-sufficient for conditional; entity lookup instead requires prompt-prefix K/V. Expression/list repair passes the in-domain gate at 8B and also requires prompt-prefix/full-cache K/V. Relation-chain repairs pass the competence gate at both 1B and 8B scale, but again require prompt-prefix/full-cache K/V. The tiny finite-state control is competent only in a one-symbol/one-step setting and also follows prompt-prefix K/V. For the 8B augmented checkpoint, latent-tail key, value, and K/V patches are ineffective across the tested cells, while prompt-prefix K/V succeeds.}
\label{tab:carrier-split}
\end{table*}

Table~\ref{tab:gpt2-block-scan} tests whether the main GPT-2 `late' split is an arbitrary boundary choice. The strongest restricted two-layer block is consistently layers 8--9, while the final 10--11 block is weak. Four-layer windows that contain 8--9 nearly recover the full late-window effect.

\begin{table*}[t]
\centering
\begingroup
\setlength{\tabcolsep}{2.4pt}
\appendixtablefont
\begin{tabular}{lccccccccc}
\toprule
\textbf{Model} & \textbf{All} & \textbf{Late} & \textbf{Nonlate} & \textbf{6--8} & \textbf{9--11} & \textbf{6--7} & \textbf{8--9} & \textbf{10--11} & \textbf{6--9} \\
\midrule
CODI-GPT2 & 1.000 & 0.994 & 0.000 & 0.211 & 0.639 & 0.000 & 0.900 & 0.011 & 0.967 \\
Sim-CoT-GPT2 & 1.000 & 1.000 & 0.161 & 0.717 & 0.956 & 0.133 & 0.983 & 0.117 & 0.989 \\
\bottomrule
\end{tabular}
\endgroup
\caption{GPT-2 contiguous layer-block scan. Entries are final-step target-win rates for oracle counterfactual value-suffix patching with cache+h, aggregated over balls/books/coins with 180 examples per row. Late is layers 8--11 and nonlate is 0--7; the remaining columns are contiguous three-layer groups, two-layer blocks, and the 6--9 sliding window.}
\label{tab:gpt2-block-scan}
\end{table*}

Table~\ref{tab:ood-numeric} checks whether the localization result is specific to the original small-integer range. The same late-value pattern appears under larger OOD numeric values. The OOD effect is strongest for CODI-GPT2, while Sim-CoT-GPT2 keeps the same direction with a larger nonlate baseline. We report only this localization result as OOD evidence; an OOD semantic-control diagnostic was directionally consistent but had a high random-source baseline, so we treat it as supplementary rather than primary evidence.

\begin{table*}[t]
\centering
\appendixtablefont
\begin{tabular}{lccccc}
\toprule
\textbf{Model / Range} & \textbf{All} & \textbf{Late} & \textbf{Late 8--9} & \textbf{Late 10--11} & \textbf{Nonlate} \\
\midrule
CODI-GPT2 / wide OOD & \twostep{0.794}{0.797} & \twostep{0.797}{0.761} & \twostep{0.719}{0.594} & \twostep{0.453}{0.133} & \twostep{0.386}{0.081} \\
Sim-CoT-GPT2 / wide OOD & \twostep{0.572}{0.692} & \twostep{0.506}{0.686} & \twostep{0.453}{0.583} & \twostep{0.183}{0.217} & \twostep{0.203}{0.217} \\
\bottomrule
\end{tabular}
\caption{OOD numeric-range localization, aggregated over balls/books/coins. Entries are target-win rates at latent steps 5/6 for oracle counterfactual value-suffix patching with cache+h. OOD examples use post-transfer and held-fixed counts in 100--500 and transfer counts in 10--99.}
\label{tab:ood-numeric}
\end{table*}

Table~\ref{tab:llama1b-stress} adds two 1B-scale diagnostics that directly address scale and hidden-state concerns. The vector-versus-cache panel shows that replacing only the current hidden state is not sufficient at 1B scale, whereas replacing the full latent cache trajectory is sufficient or near-sufficient. The wide-OOD panel shows that the all-layer value-cache transfer survives much larger numeric values on both public and matched 1B checkpoints. Because random wide-OOD sources also win often, we use the second panel only as robustness evidence, not as semantic-source separation.

\begin{table*}[t]
\centering
\begingroup
\setlength{\tabcolsep}{3pt}
\appendixtablefont
\begin{tabular}{llcccc}
\toprule
\multicolumn{6}{c}{\textbf{A. 1B vector-versus-cache diagnostic, step 6 target win}} \\
\midrule
\textbf{Model} & \textbf{Task scope} & \textbf{Hidden only} & \textbf{Latent cache} & \textbf{Cache+h} & \textbf{Last cache+h} \\
\midrule
Sim-CoT-LLaMA3-1B & balls/books/coins & 0.000 & 1.000 & 1.000 & 0.000 \\
Sim-CoT-LLaMA3-1B & + three-hop & 0.000 & 1.000 & 1.000 & 0.000 \\
CODI-LLaMA3-1B & balls/books/coins & 0.000 & 0.933 & 0.933 & 0.008 \\
CODI-LLaMA3-1B & + three-hop & 0.000 & 0.950 & 0.950 & 0.006 \\
\midrule
\multicolumn{6}{c}{\textbf{B. 1B wide numeric OOD, all-layer value cache+h, step 6 target win}} \\
\midrule
\textbf{Model} & \textbf{balls} & \textbf{books} & \textbf{coins} & \textbf{Oracle avg.} & \textbf{Random avg.} \\
\midrule
Sim-CoT-LLaMA3-1B & 0.975 & 1.000 & 1.000 & 0.992 & 0.483 \\
CODI-LLaMA3-1B & 0.975 & 1.000 & 1.000 & 0.992 & 0.475 \\
\bottomrule
\end{tabular}
\endgroup
\caption{1B calibration diagnostics. Panel A reports oracle counterfactual vector-versus-cache target-win rates. Panel B reports wide numeric OOD examples with post-transfer and held-fixed counts in 100--500 and transfer counts in 10--99. The random-source average in Panel B is high, so the wide-OOD result is used as robustness evidence for all-layer transfer rather than as a clean semantic-control result.}
\label{tab:llama1b-stress}
\end{table*}

\subsection{Component, Source, and Decoded-Behavior Details}
\label{app:component-decoded}

Tables~\ref{tab:cache-diagnostic}--\ref{tab:simcot-necessity} expand the arithmetic controls summarized in the main text: cache-only versus cache+h, hidden-only versus cache replacement, key/value separation, cross-template source controls, greedy and sampled decoding, and the weaker Sim-CoT matched-corruption replication. Table~\ref{tab:cache-diagnostic} separates late-value cache-only patching from joint cache--state patching. The final latent step is the clearest diagnostic: late-value cache-only reaches \twostep{0.469}{0.998}, whereas nonlate value cache-only remains at \twostep{0.002}{0.006}. The source hidden state is mainly needed one step earlier.

\begin{table}[t]
\centering
\appendixtablefont
\begin{tabular}{lcc}
\toprule
\textbf{Intervention} & \textbf{Target Win} & \textbf{Target Delta} \\
\midrule
Late value cache-only & \twostep{0.469}{0.998} & \twostep{12.755}{15.618} \\
Late value cache+h & \twostep{1.000}{0.998} & \twostep{15.642}{15.618} \\
Late 8--9 cache-only & \twostep{0.152}{0.887} & \twostep{9.554}{14.946} \\
Late 8--9 cache+h & \twostep{0.994}{0.887} & \twostep{15.616}{14.951} \\
Nonlate cache-only & \twostep{0.002}{0.006} & \twostep{0.774}{1.391} \\
Nonlate cache+h & \twostep{0.465}{0.006} & \twostep{12.482}{1.402} \\
\bottomrule
\end{tabular}
\caption{Cache patch diagnostic on CODI-GPT2, aggregated over balls/books/coins. Entries are latent steps 5/6.}
\label{tab:cache-diagnostic}
\end{table}

Table~\ref{tab:vector-cache} shows a direct vector-versus-cache diagnostic aggregated over CODI-GPT2 balls, books, and coins. Replacing the current hidden state alone does not explain the final-step counterfactual transfer: hidden-state-only replacement reaches \twostep{0.246}{0.000}, whereas latent-cache-only replacement reaches \twostep{0.704}{1.000}. Table~\ref{tab:simcot-vector-cache} repeats the diagnostic on Sim-CoT-GPT2 and gives the same conclusion: hidden-state-only replacement is weak at the final step, while latent-cache replacement recovers most of the effect.

\begin{table}[t]
\centering
\appendixtablefont
\begin{tabular}{lcc}
\toprule
\textbf{Intervention} & \textbf{Target Win} & \textbf{CF Gap Delta} \\
\midrule
Hidden state only & \twostep{0.246}{0.000} & \twostep{12.116}{0.003} \\
Latent cache only & \twostep{0.704}{1.000} & \twostep{19.470}{31.898} \\
Latent cache + h & \twostep{1.000}{1.000} & \twostep{31.893}{31.893} \\
Last latent cache + h & \twostep{0.267}{0.196} & \twostep{12.254}{11.521} \\
Full cache only & \twostep{0.758}{1.000} & \twostep{20.564}{32.523} \\
Full cache + h & \twostep{1.000}{1.000} & \twostep{32.522}{32.522} \\
\bottomrule
\end{tabular}
\caption{Vector-versus-cache baseline on CODI-GPT2, aggregated over balls/books/coins. Entries are latent steps 5/6.}
\label{tab:vector-cache}
\end{table}

\begin{table}[t]
\centering
\appendixtablefont
\begin{tabular}{lcc}
\toprule
\textbf{Intervention} & \textbf{Target Win} & \textbf{CF Gap Delta} \\
\midrule
Hidden state only & \twostep{0.017}{0.033} & \twostep{4.869}{6.270} \\
Latent cache only & \twostep{0.361}{0.928} & \twostep{12.806}{21.897} \\
Latent cache + h & \twostep{0.944}{1.000} & \twostep{21.961}{28.198} \\
Last latent cache + h & \twostep{0.672}{0.589} & \twostep{16.751}{15.700} \\
Full cache only & \twostep{0.983}{0.961} & \twostep{24.197}{23.052} \\
Full cache + h & \twostep{1.000}{1.000} & \twostep{29.022}{29.022} \\
\bottomrule
\end{tabular}
\caption{Vector-versus-cache baseline on Sim-CoT-GPT2, aggregated over balls/books/coins. Entries are latent steps 5/6.}
\label{tab:simcot-vector-cache}
\end{table}

Table~\ref{tab:component-sufficiency} gives a complementary component-sufficiency diagnostic. Key-only suffix patching does not transfer the counterfactual at the final latent step, even when the current hidden state is also replaced; value-only suffix patching is sufficient.

\begin{table*}[t]
\centering
\appendixtablefont
\setlength{\tabcolsep}{5pt}
\begin{tabular}{lcccc}
\toprule
\textbf{Intervention} & \textbf{TW s5} & \textbf{TW s6} & \textbf{$\Delta$ s5} & \textbf{$\Delta$ s6} \\
\midrule
Key suffix cache-only & 0.000 & 0.000 & 0.380 & 0.786 \\
Value suffix cache-only & 0.750 & 1.000 & 14.704 & 16.408 \\
KV suffix cache-only & 0.583 & 1.000 & 14.164 & 16.414 \\
Key suffix cache+h & 0.175 & 0.000 & 9.966 & 0.777 \\
Value suffix cache+h & 1.000 & 1.000 & 16.411 & 16.409 \\
KV suffix cache+h & 1.000 & 1.000 & 16.414 & 16.414 \\
\bottomrule
\end{tabular}
\caption{Key-versus-value sufficiency on CODI-GPT2 balls. TW is target win; $\Delta$ is target delta. Splitting latent steps 5 and 6 into separate columns avoids ambiguous step-pair rendering in narrow appendix columns.}
\label{tab:component-sufficiency}
\end{table*}

Table~\ref{tab:cross-template} tests whether exact counterfactual arithmetic contexts transfer across lexical domains. Cross-template exact-context sources preserve the oracle effect, while partial-variable controls remain weak even when expressed in another lexical domain.

\begin{table*}[t]
\centering
\appendixtablefont
\begin{tabular}{llcc}
\toprule
\textbf{Recipient} & \textbf{Source} & \textbf{Target Win} & \textbf{Target Delta} \\
\midrule
balls & In-template oracle & \twostep{1.000}{1.000} & \twostep{16.227}{16.206} \\
balls & Same-template partial-variable & \twostep{0.308}{0.258} & \twostep{2.751}{3.295} \\
balls & Cross-template books & \twostep{1.000}{0.992} & \twostep{16.226}{16.202} \\
balls & Cross-template coins & \twostep{1.000}{0.992} & \twostep{16.226}{16.203} \\
balls & Cross-template partial books & \twostep{0.333}{0.300} & \twostep{2.720}{3.219} \\
balls & Cross-template partial coins & \twostep{0.300}{0.267} & \twostep{3.103}{3.481} \\
balls & Random source & \twostep{0.392}{0.350} & \twostep{4.814}{5.386} \\
books & In-template oracle & \twostep{1.000}{0.992} & \twostep{15.346}{15.301} \\
books & Same-template partial-variable & \twostep{0.317}{0.292} & \twostep{2.613}{3.327} \\
books & Cross-template balls & \twostep{1.000}{0.992} & \twostep{15.347}{15.302} \\
books & Cross-template coins & \twostep{1.000}{0.992} & \twostep{15.346}{15.304} \\
books & Cross-template partial balls & \twostep{0.350}{0.283} & \twostep{1.174}{1.835} \\
books & Cross-template partial coins & \twostep{0.300}{0.283} & \twostep{1.710}{2.338} \\
books & Random source & \twostep{0.492}{0.467} & \twostep{6.642}{7.221} \\
coins & In-template oracle & \twostep{1.000}{1.000} & \twostep{16.565}{16.540} \\
coins & Same-template partial-variable & \twostep{0.342}{0.308} & \twostep{3.076}{3.586} \\
coins & Cross-template balls & \twostep{1.000}{1.000} & \twostep{16.565}{16.542} \\
coins & Cross-template books & \twostep{1.000}{1.000} & \twostep{16.565}{16.537} \\
coins & Cross-template partial balls & \twostep{0.367}{0.333} & \twostep{3.406}{3.832} \\
coins & Cross-template partial books & \twostep{0.450}{0.400} & \twostep{3.348}{3.963} \\
coins & Random source & \twostep{0.417}{0.383} & \twostep{5.999}{6.400} \\
\bottomrule
\end{tabular}
\caption{Cross-template controls on CODI-GPT2. Entries are latent steps 5/6 for late-value joint cache--state patching. Cross-template exact sources express the same counterfactual arithmetic context in another lexical domain; cross-template partial sources preserve only the partial-variable match while changing another arithmetic input.}
\label{tab:cross-template}
\end{table*}

Table~\ref{tab:simcot-cross-template} repeats the cross-template control on Sim-CoT-GPT2. Exact-context sources again track the oracle source, while partial-variable controls remain much weaker.

\begin{table}[t]
\centering
\begingroup
\setlength{\tabcolsep}{2pt}
\appendixtablefont
\begin{tabular}{@{}lcccc@{}}
\toprule
\textbf{Recip.} & \textbf{Oracle} & \textbf{Exact} & \textbf{Same Part.} & \textbf{Part.} \\
\midrule
balls & \twostep{0.983}{1.000} & \twostep{0.992}{1.000} & \twostep{0.233}{0.317} & \twostep{0.208}{0.308} \\
books & \twostep{0.967}{1.000} & \twostep{0.958}{1.000} & \twostep{0.217}{0.317} & \twostep{0.200}{0.317} \\
coins & \twostep{0.933}{0.983} & \twostep{0.933}{0.983} & \twostep{0.183}{0.283} & \twostep{0.208}{0.308} \\
\bottomrule
\end{tabular}
\endgroup
\caption{Sim-CoT-GPT2 cross-template controls. Entries are target-win rates at latent steps 5/6 for late-value cache+h patching. Exact Avg. averages the two exact cross-template source domains; Part. Avg. averages the two partial-variable cross-template controls.}
\label{tab:simcot-cross-template}
\end{table}

Table~\ref{tab:freegen} validates the main source-control pattern under greedy free decoding from the patched state. Exact-context patches decode the counterfactual answer at near-oracle rates, whereas partial-variable controls almost never do.

\begin{table*}[t]
\centering
\appendixtablefont
\begin{tabular}{llcccc}
\toprule
\textbf{Recipient} & \textbf{Source} & \textbf{N} & \textbf{Free CF} & \textbf{Free Recip.} & \textbf{Teacher Target} \\
\midrule
balls & Oracle & 60 & 0.967 & 0.017 & 0.983 \\
balls & Exact x-template avg & 120 & 0.975 & 0.008 & 0.992 \\
balls & Same partial vars & 60 & 0.000 & 0.100 & 0.367 \\
balls & Partial x-template avg & 120 & 0.000 & 0.058 & 0.292 \\
balls & Random source & 60 & 0.050 & 0.017 & 0.583 \\
books & Oracle & 60 & 0.983 & 0.000 & 1.000 \\
books & Exact x-template avg & 120 & 0.983 & 0.000 & 1.000 \\
books & Same partial vars & 60 & 0.017 & 0.083 & 0.367 \\
books & Partial x-template avg & 120 & 0.000 & 0.058 & 0.392 \\
books & Random source & 60 & 0.017 & 0.033 & 0.483 \\
coins & Oracle & 60 & 1.000 & 0.000 & 1.000 \\
coins & Exact x-template avg & 120 & 0.992 & 0.000 & 1.000 \\
coins & Same partial vars & 60 & 0.000 & 0.100 & 0.217 \\
coins & Partial x-template avg & 120 & 0.000 & 0.108 & 0.267 \\
coins & Random source & 60 & 0.033 & 0.033 & 0.417 \\
\bottomrule
\end{tabular}
\caption{Greedy free-generation validation on CODI-GPT2 at latent step 6 using late-value cache+h patching. Free CF is the fraction of patched greedy decodes whose parsed answer equals the counterfactual answer; Free Recip. is the corresponding recipient-answer rate. Teacher Target reports the teacher-forced target-win rate for the same generated state.}
\label{tab:freegen}
\end{table*}

Table~\ref{tab:freegen-simcot} summarizes the same decoded-behavior check on Sim-CoT-GPT2. Exact-context patches again decode the counterfactual answer at near-oracle rates, while partial-variable controls remain near zero.

\begin{table}[t]
\centering
\begingroup
\setlength{\tabcolsep}{1.5pt}
\appendixtablefont
\begin{tabular}{@{}lccccc@{}}
\toprule
\textbf{Recip.} & \textbf{Oracle} & \textbf{Exact} & \textbf{Same Part.} & \textbf{Part.} & \textbf{Rand.} \\
\midrule
balls & 1.000 & 1.000 & 0.000 & 0.000 & 0.033 \\
books & 0.983 & 0.983 & 0.000 & 0.008 & 0.050 \\
coins & 1.000 & 1.000 & 0.000 & 0.000 & 0.067 \\
\bottomrule
\end{tabular}
\endgroup
\caption{Sim-CoT-GPT2 greedy free-generation validation at latent step 6 using late-value cache+h patching. Entries are Free CF rates: the fraction of patched greedy decodes whose parsed answer equals the counterfactual answer.}
\label{tab:freegen-simcot}
\end{table}

Table~\ref{tab:sampled-freegen} repeats the decoded-behavior check with sampling rather than greedy decoding. The same qualitative pattern holds: oracle and exact cross-template contexts produce the counterfactual answer, while partial controls mostly produce answers outside the candidate set.

\begin{table*}[t]
\centering
\appendixtablefont
\begin{tabular}{llccccc}
\toprule
\textbf{Model} & \textbf{Source group} & \textbf{Samples} & \textbf{Free CF} & \textbf{Free Recip.} & \textbf{Out-of-set} & \textbf{Entropy} \\
\midrule
CODI-GPT2 & Oracle & 300 & 0.967 & 0.020 & 0.013 & 3.844 \\
CODI-GPT2 & Exact x-template & 600 & 0.970 & 0.022 & 0.008 & 3.849 \\
CODI-GPT2 & Same partial vars & 300 & 0.000 & 0.050 & 0.867 & 3.671 \\
CODI-GPT2 & Partial x-template & 600 & 0.000 & 0.057 & 0.848 & 3.782 \\
CODI-GPT2 & Random source & 300 & 0.017 & 0.023 & 0.843 & 3.677 \\
Sim-CoT-GPT2 & Oracle & 300 & 0.993 & 0.000 & 0.007 & 3.793 \\
Sim-CoT-GPT2 & Exact x-template & 600 & 0.995 & 0.000 & 0.005 & 3.794 \\
Sim-CoT-GPT2 & Same partial vars & 300 & 0.000 & 0.067 & 0.867 & 3.909 \\
Sim-CoT-GPT2 & Partial x-template & 600 & 0.000 & 0.077 & 0.857 & 3.765 \\
Sim-CoT-GPT2 & Random source & 300 & 0.033 & 0.100 & 0.733 & 3.513 \\
\bottomrule
\end{tabular}
\caption{Sampled free-generation diagnostic at latent step 6 with late-value cache+h patching. Each patched state is decoded with five samples using temperature 0.7, top-\emph{p} 0.95, and top-\emph{k} 40. Free CF and Free Recip. are parsed-answer rates; Out-of-set counts parsed answers outside the candidate answer set. Entropy is answer entropy in bits, averaged over task-level summaries.}
\label{tab:sampled-freegen}
\end{table*}

Table~\ref{tab:freegen-cacheonly} removes source hidden-state replacement during decoded validation. This checks whether decoded counterfactual behavior follows the transplanted cache itself, rather than only the current hidden state supplied at readout.

\begin{table*}[t]
\centering
\appendixtablefont
\begin{tabular}{llccccc}
\toprule
\textbf{Model} & \textbf{Patch} & \textbf{Oracle} & \textbf{Exact Avg.} & \textbf{Same Part.} & \textbf{Part. Avg.} & \textbf{Random} \\
\midrule
CODI-GPT2 & cache only & 0.978 & 0.978 & 0.006 & 0.006 & 0.033 \\
CODI-GPT2 & cache+h & 0.978 & 0.978 & 0.006 & 0.006 & 0.033 \\
Sim-CoT-GPT2 & cache only & 0.636 & 0.632 & 0.031 & 0.025 & 0.031 \\
Sim-CoT-GPT2 & cache+h & 0.992 & 0.992 & 0.000 & 0.003 & 0.028 \\
\bottomrule
\end{tabular}
\caption{Cache-only free-generation diagnostic. Entries are model-level averages over balls, books, and coins at latent step 6. CODI-GPT2 preserves the decoded counterfactual effect under cache-only late-value patching. Sim-CoT-GPT2 retains a substantial cache-only effect, but source hidden-state replacement improves decoded readout.}
\label{tab:freegen-cacheonly}
\end{table*}

Table~\ref{tab:simcot-necessity} is the Sim-CoT matched-corruption replication. Its selective late-value damage has the same sign as CODI-GPT2 but is weaker, so we treat it as directional necessity evidence rather than a full sufficiency-and-necessity call.

\begin{table}[t]
\centering
\appendixtablefont
\begin{tabular}{llcc}
\toprule
\textbf{Region} & \textbf{Comp.} & \textbf{Target Win} & \textbf{Rec. Drop} \\
\midrule
Late & Value & \twostep{0.975}{0.606} & \twostep{0.351}{3.912} \\
Nonlate & Value & \twostep{1.000}{1.000} & \twostep{0.001}{0.004} \\
Late 8--9 & Value & \twostep{0.986}{0.753} & \twostep{0.226}{2.353} \\
Late 10--11 & Value & \twostep{1.000}{1.000} & \twostep{0.000}{0.004} \\
Late & Key & \twostep{1.000}{1.000} & \twostep{-0.000}{0.001} \\
Nonlate & Key & \twostep{1.000}{1.000} & \twostep{0.000}{0.000} \\
\bottomrule
\end{tabular}
\caption{Sim-CoT-GPT2 necessity via matched random suffix corruption, aggregated over balls, books, and coins. Entries are latent steps 5/6.}
\label{tab:simcot-necessity}
\end{table}

\FloatBarrier
\section{Reproducibility Notes}

Code and exact command paths are publicly available at
\url{https://github.com/YIDING4869/scit-emnlp-2026}. The paper tables are
generated from JSONL outputs by the SCIT artifact builder. A public reproduction
path is: generate SCIT pairs with the arithmetic or SCIT-Bench v2 generators,
run a localization, semantic-control, competence, or carrier-split runner, and
then regenerate the markdown and CSV tables. The repository contains the code,
configs, generators, and artifact builder, but not local model checkpoints or
large generated JSONL outputs. The included evidence covers CODI-GPT2
balls/books/coins, Sim-CoT-GPT2 balls/books/coins, random-corruption necessity
results, cache diagnostics, free-generation checks, 1B/8B carrier splits,
relation-chain/finite-state repair diagnostics, and SEM summaries.

\section{Compute}

Each held-out localization or semantic diagnostic run uses one GPU. At the current sample size, a two-seed GPT-2 template run finishes within roughly tens of minutes to under one GPU-hour on a single modern academic GPU. The broader calibration package includes additional 1B/8B training and repair jobs and should be viewed as the submission evidence suite, not as the minimum cost of running \method{}.

The staged-workflow timing check uses one matched 8B prompt-carrier cell on a single H100. In two opposite execution orders, full-grid versus staged+cached wall time is $907.1\rightarrow544.6$~s and $822.9\rightarrow538.1$~s (40.0\% and 34.6\% reductions). Every corresponding Stage-1 and Stage-2 record matches the full-grid execution, with no missing, extra, or error rows. Caching 192 clean states raises peak CUDA allocation from 16.462 to 17.291~GiB. Naive two-process staging without state reuse changes wall time by only $-1.5\%$ to $+1.8\%$, so the observed speedup is attributed to staging plus caching rather than staging alone.

\end{document}